\documentclass[10pt,twocolumn,letterpaper]{article}

\usepackage[pagenumbers]{cvpr} 

\newif\ifsupp
\supptrue

\usepackage[utf8]{inputenc} 
\usepackage[T1]{fontenc}    
\usepackage{microtype}      
\usepackage{cuted}          
\usepackage{multirow}
\usepackage{adjustbox}
\usepackage{wrapfig}

\usepackage[most]{tcolorbox}
\tcbset{
  mypromptstyle/.style={
    enhanced,
    breakable,
    sharp corners,
    colback=gray!5!white,
    colframe=black,
    boxrule=0.5pt,
    width=\textwidth,
    boxsep=0pt,  
    left=0pt,
    right=0pt,
    top=0pt,
    bottom=0pt,
  }
}

\definecolor{foggyblue}{RGB}{87, 123, 193}
\newcommand{\prompttitle}[1]{%
  \noindent
  \colorbox{foggyblue}{%
    \makebox[\dimexpr\linewidth-2\fboxsep\relax][l]{%
      \textcolor{white}{\bfseries #1}
    }%
  }%
  \par\vspace{0.6em}
}

\definecolor{cvprblue}{rgb}{0.21,0.49,0.74}
\usepackage[pagebackref,breaklinks,colorlinks,allcolors=cvprblue]{hyperref}

\def\confName{CVPR}
\def\confYear{2027}

\title{FurnitureVLA: Learning Long-Horizon Bimanual Furniture Assembly with Vision-Language-Action Model}

\author{
Chenyang Ma$^{1,2}$\quad
Yue Yang$^{1,3}$\quad
Radu Corcodel$^{1}$\quad
Siddarth Jain$^{1}$\quad
Andrew Wu$^{1}$\\
Chiori Hori$^{1\dagger}$\quad
Diego Romeres$^{4\ddagger}$\\
{\normalsize $^{1}$Mitsubishi Electric Research Laboratories\quad
$^{2}$University of Oxford}\\
{\normalsize $^{3}$University of North Carolina at Chapel Hill\quad
$^{4}$Amazon}\\
{\normalsize $^{\dagger}$Corresponding author.\quad
$^{\ddagger}$Work done while at MERL.}
}

\begin{document}

\maketitle

\begin{strip}
\vspace{-3.0em}
  \centering
  \footnotesize
  \setlength{\abovecaptionskip}{0.1cm}
  \includegraphics[width=1\linewidth]{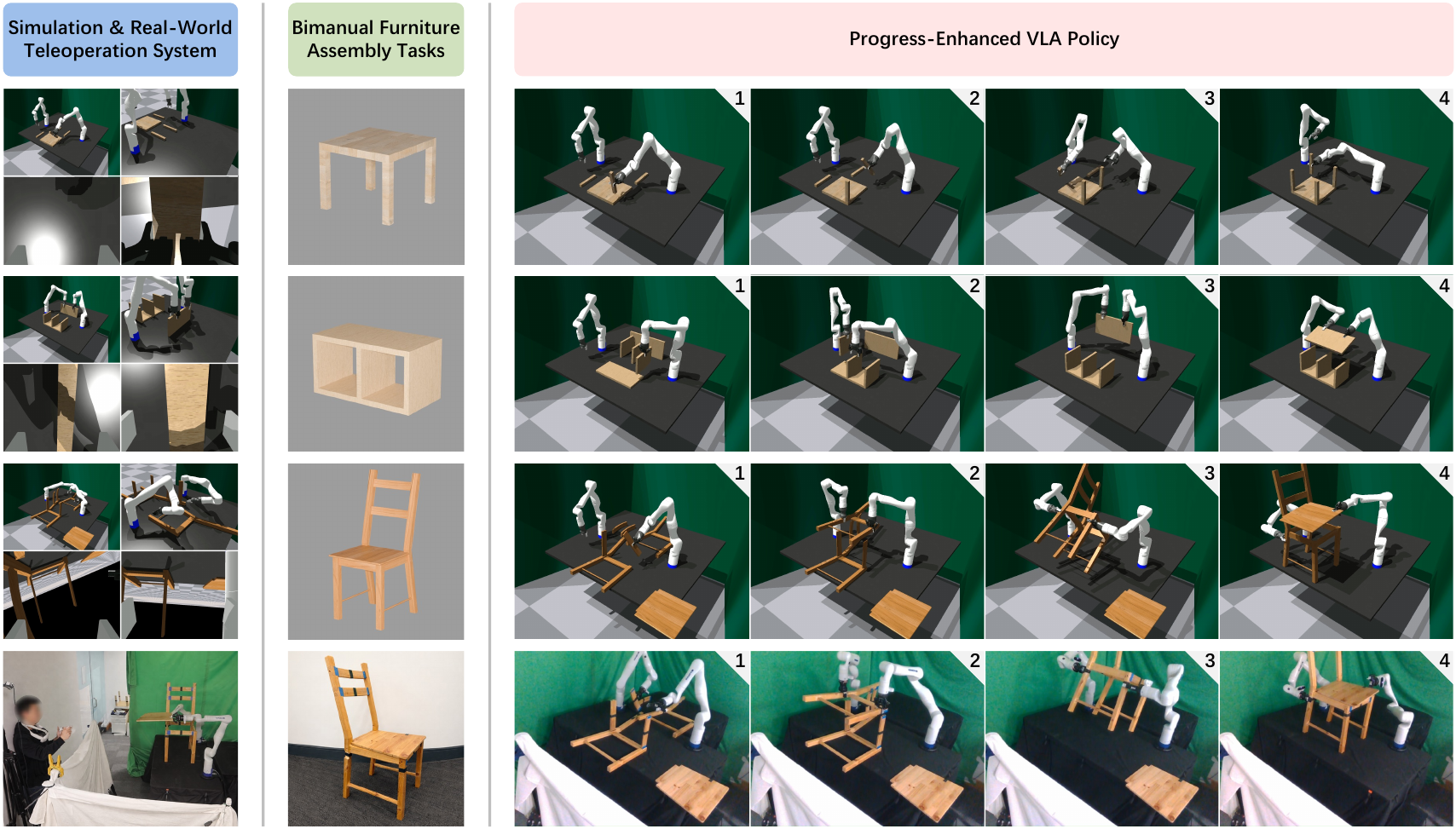}
  \vspace{-3mm}
  \captionsetup{hypcap=false}
  \captionof{figure}{\textbf{Real-scale bimanual furniture assembly with Vision-Language-Action models.} We introduce \textbf{FurnitureVLA}, the first systematic study of this challenging setting. It comprises a simulation pipeline for data generation and evaluation, and a tailored VR teleoperation system for high-quality demonstration collection. We propose a progress-enhanced VLA to tackle long-horizon assembly.}
  \captionsetup{hypcap=true}
  \label{fig:teaser}
  \vspace{0.5em}
\end{strip}

\begin{abstract}
Current work on robot furniture assembly mostly focuses on toy-scale settings or single-arm manipulation. We introduce \textbf{FurnitureVLA}, the first systematic study of real-scale bimanual furniture assembly using Vision-Language-Action models (VLAs). We formalize the task, develop a simulation pipeline for expert data generation and evaluation, and build a VR teleoperation system for single-operator collection of high-quality real-world bimanual demonstrations. To address extreme long-horizon assembly with up to 7 subtasks and 1550 control steps, we propose a progress-enhanced VLA, finetuned on semantically grounded subtasks, that jointly predicts actions and a continuous progress signal, enabling automatic subtask transitions and reducing compounding errors. Across three furniture types, FurnitureVLA raises average simulation success from 48\% for monolithically finetuned $\pi_{0.5}$ to 80\%, comparable to oracle baselines. We then ask what improves assembly success and what bounds it: perception and control design choices account for a 22\% relative gain, while physical execution, not coordination, is the main limit, with 65\% success at half the alignment tolerance. On a real Kinova Gen3 platform, the hardest IVAR chair drops only 16\% in absolute success from simulation, while finetuned $\pi_{0.5}$ reaches 0\%, and simulation co-training raises success on the LACK side table from 73\% to 93\%.
\end{abstract}

\section{Introduction}\label{sec:Introduction}
Furniture assembly is a long-standing challenge in robotics due to its long-horizon and high-precision nature. Assembling a piece of furniture requires manipulating multiple interdependent parts in strict order, where early mistakes can propagate and cause cascading failures~\cite{FurnitureBench23, LinCZ24, AnkileSS024, tian2025fabrica}. Successful assembly also demands tight geometric alignment between mating parts~\cite{ZhangWSWZT23, AnkileSSTA25, NoseworthyTWHKRFRNA25, VTRefine_huang}. These challenges are further amplified in real-scale settings: large, heavy components require sustained bimanual coordination, where two arms must jointly manipulate parts while respecting reachability limits and avoiding kinematic singularities throughout the assembly process~\cite{FuZF24, li2025robotmover}.

The goal of this work is to learn a generalist robot policy for real-scale bimanual furniture assembly using Vision-Language-Action models (VLAs), as shown in Fig.~\ref{fig:teaser}. Prior work focuses on toy-scale or scaled-down furniture with single-arm manipulation~\cite{FurnitureBench23, LinCZ24, AnkileSS024, tian2025fabrica, ZhangWSWZT23, AnkileSSTA25, NoseworthyTWHKRFRNA25, VTRefine_huang}. In this paper, we~formalize real-scale bimanual furniture assembly tasks and build a system with two complementary components. First, a bimanual simulation pipeline generates expert demonstrations via motion planning across diverse furniture types and task horizons, verifies task feasibility under reachability and kinematic constraints, and serves as a testbed for large-scale evaluation. Second, a VR teleoperation system tailored to real-scale bimanual assembly enables coordinated dual-arm control by a single teleoperator on a Kinova Gen3 platform for high-quality demonstration collection.

With this system in place, directly finetuning a monolithic VLA policy on full-length bimanual demonstrations remains suboptimal, as current VLA models are most effective on short-horizon tasks with relaxed alignment requirements~\cite{KimPKXB0RFSVKBT24, generalist2025gen0}. Long-horizon trajectories induce distribution drift, where small deviations can drive the policy into out-of-distribution states and lead to compounding errors~\cite{RossGB11, YangZDBS25}. To address this, we decompose each assembly into semantically grounded subtasks and finetune the VLA on shorter, language-conditioned segments. This narrows the within-subtask state distribution and provides cues for task stages. Importantly, we place subtask boundaries at stable, contact-free post-retreat states rather than at the contact-rich states right after part assembly, so small execution errors do not amplify and each subtask starts from a consistent state distribution. However, this requires determining when to transition between subtasks during execution. We address this by introducing a progress-enhanced VLA that jointly predicts actions and an explicit subtask progress signal to trigger transitions during inference.

Finally, subtask decomposition with learned transitions reduces compounding error, but long-horizon assembly can still fail during execution, and it is unclear which factors set the limit. We therefore ask what improves assembly success, via perception and control design choices, and what bounds it, via alignment tolerance, failure analysis, and demonstration quantity. We evaluate in simulation and on a real bimanual Kinova Gen3 platform, where we also study co-training with simulation data. In summary, we introduce \textbf{FurnitureVLA}, with the following contributions:

\begin{itemize}
\renewcommand\labelitemi{\scalebox{0.75}{$\bullet$}}
\item The first systematic study of VLAs for real-scale bimanual furniture assembly, with a simulation pipeline for expert data generation and large-scale evaluation and a VR teleoperation system for high-quality real-world demonstrations.
\item A progress-enhanced VLA that jointly predicts actions and subtask progress for automatic subtask transitions, raising average simulation success from 48\% for monolithic finetuned $\pi_{0.5}$ to 80\%, comparable to oracle baselines.
\item An analysis of what improves and what bounds assembly success: perception and control design choices give a 22\% relative gain, while failure analysis and alignment-tolerance sweep show physical execution, not coordination, is the main limit, with 65\% success at half tolerance.
\item Real-world validation on a bimanual Kinova Gen3 platform, where the hardest IVAR chair drops only 16\% in absolute success from simulation, while monolithic finetuned $\pi_{0.5}$ reaches 0\%, and simulation co-training raises success on the LACK side table from 73\% to 93\%.
\end{itemize}

\begin{figure*}[t]
  \centering
  \footnotesize
  \setlength{\abovecaptionskip}{0.1cm}
  \includegraphics[width=1\linewidth]{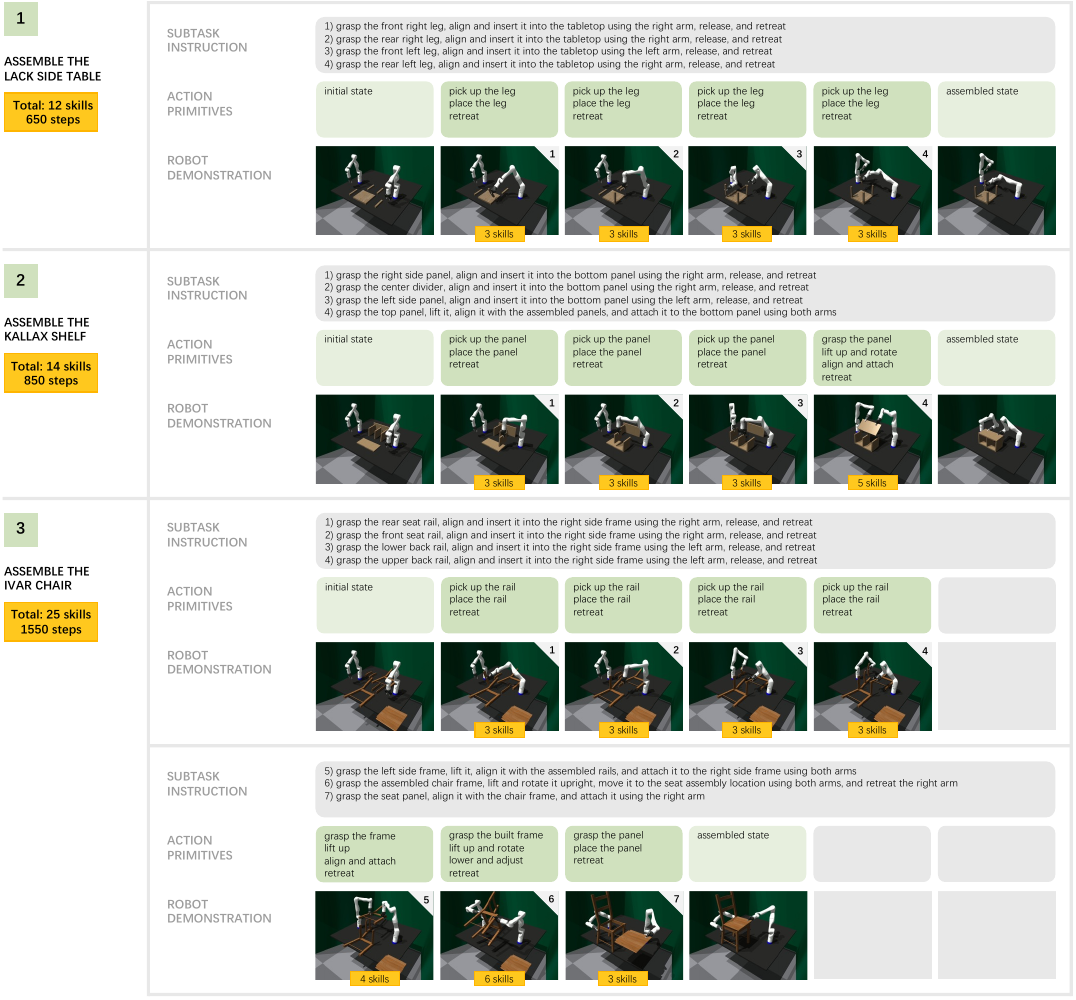}
  \vspace{-3mm} \caption{\textbf{Bimanual furniture assembly tasks.} These tasks require executing diverse manipulation skills over extremely long horizons. A simple item (LACK side table) requires 12 skill executions (650 steps), while a complex assembly (IVAR chair) requires 25 (1550 steps).}
  \label{fig:tasks}
\end{figure*}

\section{Related Work}\label{sec:Related Work}

\noindent\textbf{Robot Furniture Assembly.} Early work relies on engineered perception and task-and-motion planning~\cite{ZhouP18}; later work translates assembly manuals into plans with learned contact corrections and vision-language alignment~\cite{tie2025manual, ma2024spatialpin, tian2025fabrica}. On simulation benchmarks~\cite{LeeHL21, FurnitureBench23}, learning-based methods study large-scale imitation~\cite{AnkileSS024}, residual policies for precision~\cite{AnkileSSTA25}, sim-to-real transfer from limited real data~\cite{TangLAHS0FN23, ZhangWSWZT23, VTRefine_huang}, and human-robot collaboration~\cite{GiacomuzzoTJR24}. Most target single-arm, task-specific policies at toy scale, leaving open generalist policies for real-scale bimanual assembly.

\vspace{1.0em}
\noindent\textbf{VLAs for Long-Horizon Manipulation.} Long-horizon execution is brittle under error accumulation and covariate shift~\cite{RossGB11, sun2024arch, YangZDBS25, ma2025coopera}, so demonstrations are commonly split into phases: modular skills with external switching~\cite{fan2025longvla, seqvla_yang, yang2026lilo, lu2026pgx}, modules that replan and recover under a detector~\cite{lu2025kitchenvla, ma2026cyclevla}, or VLAs that reason about the next subtask~\cite{ZawalskiCPMFL24, pi0_5_pi, ShiIEKPVTWWFLDG25}, at higher computational cost. Our progress-enhanced VLA switches subtasks without an external stage estimator.

\vspace{1.0em}
\noindent\textbf{Precision Limits of Generalist Policies.} Generalist policies such as VLAs~\cite{KimPKXB0RFSVKBT24, generalist2025gen0, gr00t_fan, pi0_5_pi} remain precision-limited: cross-embodiment standardization can coarsen action representations~\cite{ONeillRMGPLPGMJ24} and discretized action tokenization can fail on dexterous, high-frequency data~\cite{pertsch2025fast}, as seen in real-robot benchmarks~\cite{atreya2025roboarena}. ACT~\cite{aloha23} reduces compounding error with action chunking and temporal ensembling, and Openpi Comet~\cite{bai2025openpi} shows that perception and control-loop choices affect accuracy. We study these in real-scale bimanual assembly, asking which improve success and what bounds~it.

\section{Tasks and System Setup}
We first formalize the task and its assumptions (Sec.~\ref{sec:Bimanual Furniture Assembly Tasks}), then describe the simulation pipeline (Sec.~\ref{sec:Simulation Data Generation and Success Criteria}) and the VR teleoperation system (Sec.~\ref{sec:Real-World Data Collection via VR Teleoperation}).

\subsection{Bimanual Furniture Assembly Tasks}\label{sec:Bimanual Furniture Assembly Tasks}
\noindent\textbf{Hardware Setup.} Assembly is performed on a tabletop with two Kinova Gen3 7-DoF arms, a Robotiq Hand-E gripper on the left and a 2F-85 on the right (Fig.~\ref{fig:system}). Beyond the front and wrist cameras common in prior VLA work~\cite{pi0_5_pi}, we add a rear camera, as large parts and frequent bimanual interactions often occlude the frontal view.

\begin{wrapfigure}{R}{0.25\textwidth}
    \centering
    \vspace{-1.3em}
    \includegraphics[width=\linewidth]{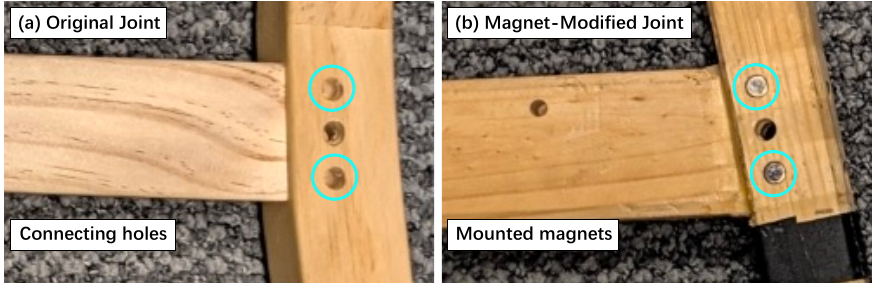}
    \vspace{-2.0em} \caption{\textbf{What the magnets replace.} An IVAR joint with (a) its original fastener and (b) magnets at the connection points. The holes still dictate where the parts must meet; the magnets only replace the tool-and-torque fastening step.}
    \label{fig:magnet}
\vspace{-1.0em}
\end{wrapfigure}

\noindent\textbf{Assumptions.} We make two assumptions. First, we replace the original fasteners with magnets mounted at the connecting holes of each joint (Fig.~\ref{fig:magnet}), so that parts attach once aligned. The magnets replace the fastening operation, not the alignment: the connecting holes still define where the parts must meet, and the policy must bring the mating points within the magnets' capture range (about 1\,cm, Fig.~\ref{fig:eval}) before the joint seats. This range is measured on the real joint rather than chosen in simulation, and matches the 1--2~cm pre-insertion alignment assumed in insertion work such as IndustReal~\cite{TangLAHS0FN23}. What we bypass is screwing, which requires tool use and tightening torque while both arms support large, heavy parts; this is an orthogonal and harder problem~\cite{HanCSJLCPB22, TangJ23}, is difficult to simulate, and is likewise excluded by prior assembly benchmarks~\cite{LeeHL21, FurnitureBench23}. To ensure our results do not hinge on this capture range, Sec.~\ref{sec:What Bounds Assembly Success} evaluates the policy under progressively tighter alignment tolerances, down to the millimeter level. Second, each furniture part is randomly initialized within 3\,cm in position and 5$^\circ$ in orientation of its nominal pose, consistent with common VLA settings~\cite{KimPKXB0RFSVKBT24, gr00t_fan, pi0_5_pi}.

\setlength{\columnsep}{3pt}
\setlength{\columnsep}{10pt}

\vspace{0.5em}
\noindent\textbf{Task Properties.}
We use three IKEA items of increasing difficulty, the LACK side table, KALLAX shelf, and IVAR chair, each decomposed into subtasks with concise, semantically grounded language instructions (Fig.~\ref{fig:tasks}). Four properties make these tasks challenging:

\vspace{0.5em}
\noindent\textbf{1) Long-horizon execution.} Each assembly requires completing a strict sequence of subtasks, with horizons ranging from 4 (LACK, KALLAX) to 7 (IVAR), and 650--1550 control steps (65--155 seconds), exceeding prior benchmarks~\cite{MeesHRB22, LiuZGFLZS23, robocasa2024}.

\vspace{0.5em}
\noindent\textbf{2) Tolerance-bounded insertion.} Each subtask requires bringing mating parts within the alignment tolerance, 1--2\,cm in translation and about $4^\circ$ in rotation (success criteria in Sec.~\ref{sec:Simulation Data Generation and Success Criteria}), which is more demanding relative to part size than in prior toy-scale assembly benchmarks~\cite{LeeHL21, FurnitureBench23}.

\vspace{0.5em}
\noindent\textbf{3) Diverse manipulation skills.} Each subtask is specified by a structured language instruction (e.g., ``grasp, align, and insert a part'') and requires composing multiple skills, such as grasping, alignment, insertion, lifting, and rotation. A simple assembly (LACK, 4 subtasks) involves roughly 12 skill executions, while IVAR involves 25 (complete set in Appendix~\ref{sec:Simulation Implementation Details}) due to its longer horizon and additional bimanual operations such as coordinated lifting and rotation of assembled structures. These skills must be performed under occlusion, contact-rich insertion, and collision-aware maneuvering around partially assembled structures.

\vspace{0.5em}
\noindent\textbf{4) Bimanual coordination.} Several subtasks require both arms to act in concert over large, heavy parts. For example, attaching the KALLAX top panel requires simultaneously lifting, rotating, and aligning a full-width panel with both arms, and assembling IVAR requires lifting and rotating the partially assembled chair frame upright before attaching the seat panel. These bimanual subtasks involve up to six sequential action primitives and must maintain reachability and avoid singularities throughout.

\subsection{Simulation Data and Success Criteria}\label{sec:Simulation Data Generation and Success Criteria}
\noindent\textbf{Data Generation via Motion Planning.} For single-arm actions, we plan end-effector pose trajectories directly. For bimanual actions, we disable physics and plan the trajectory of the manipulated object, with both arms following as rigidly attached end-effectors. All trajectories are validated by replay in simulation. Initial part placements are designed for feasibility: parts remain within reach, avoid inter-object collisions, and stay clear of kinematic singularities.

\vspace{0.5em}
\noindent\textbf{Assembly Success Criteria.} As in FurnitureBench~\cite{FurnitureBench23}, we designate one part of each furniture as the base and evaluate every other part by its pose relative to it. With $\mathbf{T}_b, \mathbf{T}_k \in SE(3)$ the poses of the base and the $k$-th part, $\mathbf{T}_{\text{rel}} = \mathbf{T}_b^{-1} \mathbf{T}_k$ is compared against the ground-truth assembled pose $\mathbf{T}_{\text{rel}}^{*}$. A part is assembled if:
\begin{equation}
\begin{aligned}
\lvert t_{\text{rel},i} - t_{\text{rel},i}^{*} \rvert &\leq \epsilon, && i \in \{x,y,z\}, \\
\cos\bigl(\mathbf{r}_{\text{rel},j}, \mathbf{r}_{\text{rel},j}^{*}\bigr) &\geq \delta, && j \in \{u,v,w\},
\end{aligned}
\label{eq:2}
\end{equation}
where $t_{\text{rel},i}$ is the $i$-th component of the translation and $\mathbf{r}_{\text{rel},j}$ the $j$-th column of the rotation matrix of $\mathbf{T}_{\text{rel}}$. 

We set $\epsilon = 1\,\text{cm}$ for small parts and $2\,\text{cm}$ for large parts, and $\delta = 0.998$ (${\approx}4.0^\circ$ per rotation axis), the same alignment tolerance level as FurnitureBench~\cite{FurnitureBench23} and IndustReal~\cite{TangLAHS0FN23}, but applied to much larger parts, hence more demanding relative to part size. A full assembly succeeds only when all parts satisfy both criteria.

\subsection{Real-World Data via VR Teleoperation}
\label{sec:Real-World Data Collection via VR Teleoperation}

\begin{figure}[t]
  \centering
  \footnotesize
  \setlength{\abovecaptionskip}{0.1cm}
  \includegraphics[width=1\linewidth]{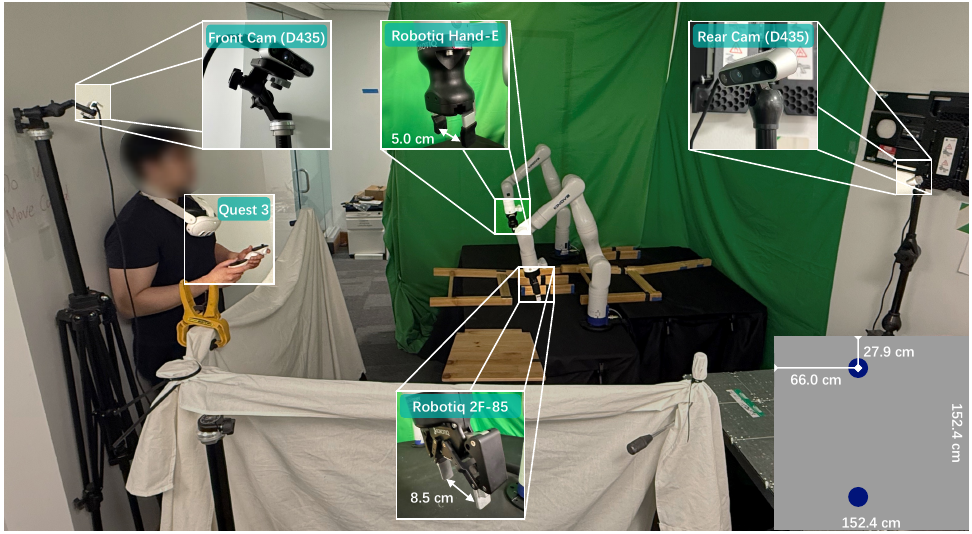}
  \vspace{-3mm} \caption{\textbf{System setup.} The teleoperator wears a Meta Quest 3 headset at the neck to track hand poses. A green screen reduces visual noise; a white cover occludes the teleoperator. Bottom-right: tabletop schematic with symmetric dual Kinova placement.}
  \label{fig:system}
  \vspace{-1.0em} 
\end{figure}

Our VR teleoperation system (Fig.~\ref{fig:system}) enables coordinated dual-arm control by a single teleoperator, with three design principles tailored for long-horizon bimanual assembly:

\vspace{0.5em}
\noindent\textbf{1) Decoupled translation and rotation control.} Translation is controlled on all three axes simultaneously for smooth end-effector motion, while rotation is handled independently to avoid unintended orientation changes, which matters for insertion subtasks such as aligning IVAR rails, where a misaligned rail fails to attach.

\vspace{0.5em}
\noindent\textbf{2) Pre-defined grasp primitives.} Button-triggered grasp poses with discrete $90^\circ$ orientation variants snap the end-effector orientation to a preset while keeping its current position, removing manual wrist rotation and enforcing consistent, task-specific grasps of slender rails, flat seat panels, and large side frames.

\vspace{0.5em}
\noindent\textbf{3) Synchronized bimanual control.}
In synchronized mode both arms execute mirrored commands simultaneously, enabling efficient repositioning, alignment, and rotation of large and heavy furniture components. This is critical for subtasks such as jointly lifting and rotating the IVAR chair frame, where coordinated motion is required to maintain stability and achieve accurate placement.


\section{FurnitureVLA: Progress-Enhanced VLA}
We propose a progress-enhanced VLA finetuned on semantically grounded subtasks to mitigate distribution drift in long-horizon tasks, jointly predicting actions and a progress signal to trigger subtask transitions (Fig.~\ref{fig:pipeline}). We build on a VLA policy $\pi_\theta(a_{t:t+H-1} \mid o_t, g)$ that maps an observation $o_t$ (RGB images and proprioception) and a language instruction $g$ to a chunk of $H$ future actions via flow matching~\cite{pi0_5_pi}, where $a_t \in \mathbb{R}^{14}$ concatenates the absolute end-effector pose and gripper state of both arms. We first describe continuous progress assignment using action primitives with post-retreat subtask boundaries (Sec.~\ref{sec:Progress-Enhanced Subtask Finetuning}), then how it enables subtask transitions during inference (Sec.~\ref{sec:Inference and Subtask Transition}).

\subsection{Progress-Enhanced Subtask Finetuning}
\label{sec:Progress-Enhanced Subtask Finetuning}
\begin{figure}[t]
  \centering
  \footnotesize
  \setlength{\abovecaptionskip}{0.1cm}
  \includegraphics[width=1\linewidth]{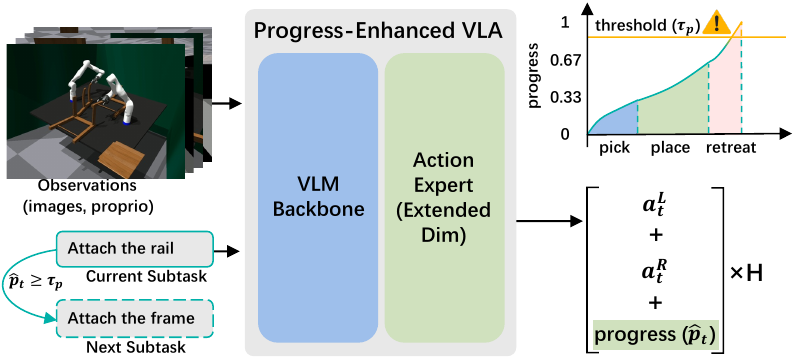}
  \vspace{-3mm} \caption{\textbf{FurnitureVLA: progress-enhanced VLA.} Each assembly is decomposed into subtasks with continuous progress labels from action primitives. The VLA jointly predicts actions and a progress signal, which triggers subtask transitions at inference.}
  \label{fig:pipeline}
  \vspace{-1.0em} 
\end{figure}

\noindent\textbf{Progress Assignment via Action Primitives.}
We decompose the task $g$ into subtasks $G = (g_1, \ldots, g_K)$ and augment the bimanual action $a_t$ with a scalar progress signal $p_t$, giving $\tilde{a}_t = [a_t^\top, p_t]^\top \in \mathbb{R}^{15}$. Each subtask consists of $N_k$ action primitives (Fig.~\ref{fig:tasks}), such as pick up, place, and retreat, each defining a fixed stage of completion (Fig.~\ref{fig:pipeline}). With $s_i$ the start timestep of the $i$-th primitive and $s_{N_k}$ the end of the subtask, segment $i \in \{0, \ldots, N_k-1\}$ occupies the uniform interval $\bigl[\tfrac{i}{N_k}, \tfrac{i+1}{N_k}\bigr]$, and progress is:
\begin{equation}
p_t = \frac{i}{N_k} + \frac{1}{N_k} \cdot \frac{t - s_i}{s_{i+1} - s_i}, 
\quad s_i \le t < s_{i+1}.
\label{eq:3}
\end{equation}
This yields a monotonic signal from $0$ to $1$ over each subtask; since each segment is a fixed control objective with smoothly evolving motion, linear interpolation is a natural time-parameterization of progress. We finetune the policy via flow matching on augmented action chunks, $\pi_\theta(\tilde{a}_{t:t+H-1} \mid o_t, g_k)$, to jointly predict actions and progress.

\vspace{0.5em}
\noindent\textbf{Post-Retreat Subtask Boundaries.} We define subtask boundaries after retreat rather than immediately after assembly. Transitioning at assembly completion is fragile: small deviations can leave an insertion incomplete or a contact unstable, and contact-rich post-assembly states amplify execution errors into large state variations across rollouts, widening the initial-state distribution of the next subtask. Post-retreat states are free of contact and force constraints, so small errors do not amplify, yielding a narrower initial-state distribution and less cross-subtask distribution shift.

\subsection{Inference and Subtask Transition}
\label{sec:Inference and Subtask Transition}
At inference time, the policy predicts $\tilde{a}_{t:t+H-1}$ conditioned on the current subtask $g_k$. The predicted progress $\hat{p}_t$ triggers a transition when it exceeds a threshold $\tau_p$; to suppress spurious predictions, let $h_t = (\hat{p}_t \ge \tau_p)$ and trigger only if:
\begin{equation}
\text{TRANSIT} =
\begin{cases}
1, & h_t \wedge h_{t-1}, \\
1, & h_t \wedge \neg h_{t-1} \wedge \neg h_{t-2} \wedge \exists\,\Delta \ge 3:\ h_{t-\Delta}, \\
0, & \text{otherwise.}
\end{cases}
\label{eq:4}
\end{equation}
This filters isolated spikes while remaining responsive to persistent signals. Upon confirmation, the system advances to $g_{k+1}$, resets progress, and clears action buffers.


\section{Simulation Experiments}
We examine: 1) whether the progress-enhanced VLA improves long-horizon execution over monolithic learned policies and conventional subtask coordination; 2) if progress-driven transitions are reliable, via threshold and design ablations; and 3) what improves assembly success, via perception and control design factors, and what bounds it, via alignment tolerance, failure breakdown, and demonstration quantity.

\begin{table}[t]
\setlength{\abovecaptionskip}{0.1cm}
    \footnotesize
    \caption{\textbf{Assembly performance against baselines} (success rates~$\uparrow$). All baselines use the same inference settings as FurnitureVLA. Best method without privileged information in bold.}
    \begin{adjustbox}{width=1.0\linewidth,center}
    \centering
    \begin{tabular}{llcccc}
    \toprule
    \textbf{Category} & \textbf{Method} & \textbf{LACK} & \textbf{KALLAX} & \textbf{IVAR} & \textbf{Average} \\ \noalign{\vskip 0.3ex}
    \hline \noalign{\vskip 1.0ex}
    \textbf{Planning} & Scripted motion planner$^\dagger$ & 0.99 & 0.62 & 0.37 & 0.66 \\
    \noalign{\vskip 0.3ex}
    \midrule \noalign{\vskip 0.2ex}
    \multirow{6}{*}[-0.75ex]{\textbf{Learning}} & Diffusion Policy~\cite{chi2023diffusion} (per furniture) & 0.20 & 0.18 & 0.06 & 0.15 \\
    \noalign{\vskip 0.5ex}
    & Diffusion Policy (one for three) & 0.38 & 0.40 & 0.14 & 0.31 \\
    \noalign{\vskip 0.5ex}
    & ACT~\cite{aloha23} (per furniture) & 0.41 & 0.51 & 0.11 & 0.34 \\
    \noalign{\vskip 0.5ex}
    & ACT (one for three) & 0.43 & 0.39 & 0.17 & 0.33 \\
    \noalign{\vskip 0.5ex}
    & $\pi_{0.5}$~\cite{pi0_5_pi} (zero-shot) & 0.00 & 0.00 & 0.00 & 0.00 \\
    \noalign{\vskip 0.5ex}
    & $\pi_{0.5}$ (monolithic finetuned) & 0.91 & 0.11 & 0.41 & 0.48 \\
    \noalign{\vskip 0.3ex}
    \midrule \noalign{\vskip 0.2ex}
    \multirow{3}{*}[-0.5ex]{\textbf{Hybrid}} & Hi Robot-style~\cite{ShiIEKPVTWWFLDG25} & 0.07 & 0.12 & 0.00 & 0.06 \\
    \noalign{\vskip 0.5ex}
    & Fabrica-style$^\dagger$~\cite{tian2025fabrica} & 0.94 & 0.68 & 0.57 & 0.73 \\
    \noalign{\vskip 0.5ex}
    & Oracle transitions$^\dagger$ & 0.96 & 0.86 & 0.54 & 0.79 \\
    \noalign{\vskip 0.3ex} 
    \midrule \noalign{\vskip 0.2ex}
    & \textbf{FurnitureVLA (ours)} & \textbf{0.98} & \textbf{0.85} & \textbf{0.56} & \textbf{0.80} \\
    \bottomrule
    \end{tabular}
    \label{tab:long_horizon}
\end{adjustbox}
{\noindent\scriptsize $^\dagger$Privileged simulator information: ground-truth part poses (motion planner) or oracle subtask transitions (hybrid methods).\par}
\vspace{-1.5em}
\end{table}

\subsection{Long-Horizon Assembly Performance}\label{sec:Long-Horizon Assembly Performance}

\noindent\textbf{Setup.} We use Isaac Gym~\cite{MakoviychukWGLS21} and finetune one $\pi_{0.5}$~\cite{pi0_5_pi} model on all furniture with 500 demonstrations each, with $\tau_p = 0.95$ at inference and the default design factors of Sec.~\ref{sec:Design Factors to Improve Assembly Success}. We evaluate 100 rollouts per furniture and report full-assembly success; a rollout terminates at the first failed subtask, so per-subtask success decreases monotonically. See Appendices~\ref{sec:Additional Details of Simulation} and~\ref{sec:FurnitureVLA Details} for implementation details.

\setlength{\columnsep}{3pt}
\begin{wrapfigure}{R}{0.23\textwidth}
    \centering
    \vspace{-1.2em}
    \includegraphics[width=\linewidth]{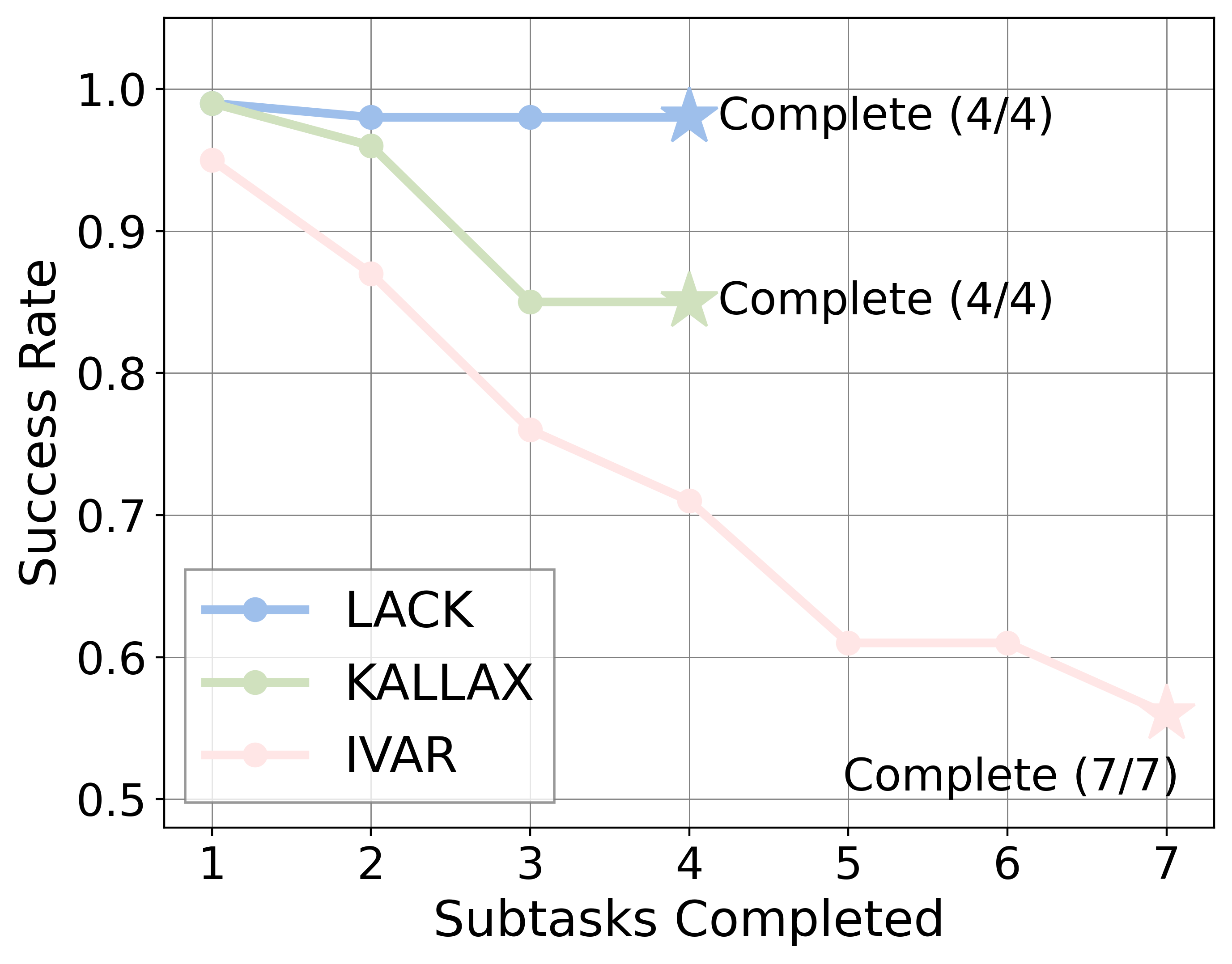}
    \vspace{-2.0em} \caption{\textbf{Subtask success rates} (monotonically decreasing).}
    \label{fig:main_results}
\vspace{-1.0em}
\end{wrapfigure}
\setlength{\columnsep}{10pt}

\vspace{0.5em}
\noindent\textbf{Baselines.} \textbf{1) Learned policies without a VLA backbone}: Diffusion Policy and ACT, trained per furniture or jointly. \textbf{2) VLAs without decomposition}: zero-shot and monolithically finetuned $\pi_{0.5}$ with one global instruction (e.g., ``assemble the IVAR chair''). \textbf{3) Planning-based and hybrid pipelines}: the data-collection motion planner with ground-truth part poses, replayed open-loop in physics, and three coordinators that use our finetuned VLA as the low-level controller: a Hi Robot-style~\cite{ShiIEKPVTWWFLDG25} VLM (GPT-6 Astra~\cite{GPT-6-astra_openai}) that judges subtask completion from images at 1\,Hz, a Fabrica-style$^\dagger$~\cite{tian2025fabrica} hybrid where the planner handles grasping and transport and the VLA only the contact-rich insertion, and oracle transitions$^\dagger$ switched by the simulator. Since decomposition is known to benefit long-horizon manipulation~\cite{pi0_5_pi, ShiIEKPVTWWFLDG25}, the hybrids hold the VLA fixed to isolate learned progress from conventional coordination; the $\dagger$ coordinators use a privileged assembly predicate unavailable on real robots and upper-bound perfect completion sensing.

\begin{table*}[t]
\setlength{\abovecaptionskip}{0.1cm}
    \footnotesize
    \caption{\textbf{Impact of perception and control design factors} on assembly success (success rates~$\uparrow$). Each column varies one factor from our default ($\lambda{=}-0.1$, horizon 10 or 25 for KALLAX, full views, 448 px; bold), a 22\% relative gain over the mean of all ablated settings (0.65).}
    \begin{adjustbox}{width=1.0\linewidth,center}
    \centering
    \begin{tabular}{lccccccccccccccc}
    \toprule
     \multirow{2}{*}[-0.5ex]{\textbf{Furniture}} & \multicolumn{6}{c}{\textbf{Temporal Ensembling}} & \multicolumn{3}{c}{\textbf{Action Horizon}} & \multicolumn{3}{c}{\textbf{Viewpoint}} & \multicolumn{3}{c}{\textbf{Resolution}} \\
     \cmidrule(lr){2-7} \cmidrule(lr){8-10} \cmidrule(lr){11-13} \cmidrule(lr){14-16}
     & n/a & -0.25 & -0.1 & 0.0 & 0.1 & 0.25 & 5 & 10 & 25 & full & depth & w/o rear & 224 & 300 & 448 \\ \noalign{\vskip 0.3ex}
    \hline \noalign{\vskip 1.0ex}
    \textbf{LACK}   & 0.76 & 0.90 & \textbf{0.98} & 0.91 & 0.91 & 0.92 & 0.92 & \textbf{0.98} & 0.75 & \textbf{0.98} & 0.69 & 0.45 & 0.59 & 0.97 & \textbf{0.98} \\
    \noalign{\vskip 0.5ex}
    \textbf{KALLAX} & 0.83 & 0.81 & \textbf{0.85} & 0.44 & 0.84 & 0.80 & 0.43 & 0.67 & \textbf{0.85} & \textbf{0.85} & 0.45 & 0.57 & 0.77 & 0.68 & \textbf{0.85} \\
    \noalign{\vskip 0.5ex}
    \textbf{IVAR}   & 0.36 & 0.53 & \textbf{0.56} & 0.52 & 0.55 & 0.54 & 0.44 & \textbf{0.56} & 0.41 & \textbf{0.56} & 0.48 & 0.34 & 0.43 & 0.50 & \textbf{0.56} \\
    \noalign{\vskip 0.5ex}
    \textbf{Average} & 0.65 & 0.75 & \textbf{0.80} & 0.62 & 0.77 & 0.75 & 0.60 & \textbf{0.74} & 0.67 & \textbf{0.80} & 0.54 & 0.45 & 0.60 & 0.72 & \textbf{0.80} \\
    \bottomrule
    \end{tabular}
    \label{tab:design_factors}
\end{adjustbox}
\vspace{-0.5em}
\end{table*}

\vspace{0.5em}
\noindent\textbf{Analysis.} Zero-shot $\pi_{0.5}$ fails on all furniture, and FurnitureVLA consistently outperforms monolithic finetuning, most on KALLAX, whose large, heavy parts make execution sensitive to accumulated errors (Table~\ref{tab:long_horizon}). Fig.~\ref{fig:main_results} breaks success down by subtask: IVAR failures concentrate on the fifth, where both arms lift and attach the left side frame to the partial assembly. Even the motion planner, with ground-truth poses, drops substantially on KALLAX and IVAR under physics replay, underscoring the difficulty of reliable physical execution. FurnitureVLA exceeds both Hi Robot-style and the privileged Fabrica-style pipeline and closely matches oracle transitions, supporting learned progress as a substitute for privileged completion sensing; the small gap to the oracle suggests that physical execution, rather than coordination, is the main remaining limit (Sec.~\ref{sec:What Bounds Assembly Success}).

\subsection{Analysis of Progress-Driven Transitions}\label{sec:Analysis of Progress-Driven Transitions}
\vspace{-0.25em}

\begin{table}[t]
\setlength{\abovecaptionskip}{0.1cm}
    \footnotesize
    \caption{\textbf{Progress threshold sensitivity and alternative designs} (success rates~$\uparrow$). Default FurnitureVLA (learned continuous progress, post-retreat boundaries, $\tau_p{=}0.95$ and Eq.~\ref{eq:4}) filter in bold.}
    \begin{adjustbox}{width=1.0\linewidth,center}
    \centering
    \begin{tabular}{llcccc}
    \toprule
    \textbf{Ablation} & \textbf{Method} & \textbf{LACK} & \textbf{KALLAX} & \textbf{IVAR} & \textbf{Average} \\ \noalign{\vskip 0.3ex}
    \hline \noalign{\vskip 1.0ex}
    \multirow{3}{*}[-0.5ex]{\textbf{Design}} & Fixed duration from demonstrations & 0.72 & 0.56 & 0.10 & 0.46 \\
    \noalign{\vskip 0.5ex}
    & SeqVLA-style~\cite{seqvla_yang} binary head & 0.66 & 0.70 & 0.23 & 0.53 \\
    \noalign{\vskip 0.5ex}
    & Boundaries at completion ($\tau_p{=}0.95$) & 0.60 & 0.37 & 0.25 & 0.41 \\
    \noalign{\vskip 0.5ex}
    & Boundaries at completion ($\tau_p{=}0.98$) & 0.79 & 0.71 & 0.32 & 0.61 \\
    \noalign{\vskip 0.3ex}
    \midrule \noalign{\vskip 0.2ex}
    \multirow{5}{*}[-1.0ex]{\textbf{Threshold}} & $\tau_p{=}0.80$ & 0.47 & 0.78 & 0.39 & 0.55 \\
    \noalign{\vskip 0.5ex}
    & $\tau_p{=}0.85$ & 0.65 & 0.66 & 0.40 & 0.57 \\
    \noalign{\vskip 0.5ex}
    & $\tau_p{=}0.90$ & 0.95 & 0.87 & 0.52 & 0.78 \\
    \noalign{\vskip 0.5ex}
    & $\tau_p{=}0.98$ & 0.89 & 0.85 & 0.58 & 0.77 \\
    \noalign{\vskip 0.5ex}
    & $\tau_p{=}0.95$ without filter & 0.97 & 0.81 & 0.49 & 0.76 \\
    \noalign{\vskip 0.3ex}
    \midrule \noalign{\vskip 0.2ex}
    & \textbf{FurnitureVLA (continuous, $\tau_p{=}0.95$)} & \textbf{0.98} & \textbf{0.85} & \textbf{0.56} & \textbf{0.80} \\
    \bottomrule
    \end{tabular}
    \label{tab:progress}
\end{adjustbox}
\vspace{-1.0em}
\end{table}

FurnitureVLA switches subtasks only on its own progress estimate. Sec.~\ref{sec:Long-Horizon Assembly Performance} shows that this matches oracle transitions at the default configuration; we now test its sensitivity to the threshold and filter, and if a learned, continuous progress signal is needed at all. Appendix~\ref{sec:Analyses Details} further reports the offline accuracy of the progress head and a discrete-progress variant.

\vspace{0.5em}
\noindent\textbf{Progress Threshold Sensitivity.}
We vary $\tau_p$ from 0.80 to 0.98 and ablate the confirmation filter (Table~\ref{tab:progress}). Success is stable for $\tau_p \geq 0.90$, and removing the filter lowers success on all furniture, confirming it suppresses spurious switches.

\vspace{0.5em}
\noindent\textbf{Alternative Designs.}
We test \textbf{1) fixed-duration switching}, setting each subtask's duration to the 90th percentile of its demonstration lengths; \textbf{2) SeqVLA-style binary head}~\cite{seqvla_yang}, BCE-trained on action-expert features, labeled 1 where progress (Eq.~\ref{eq:3}) ${\geq} 0.95$, else 0, with the same filter at threshold $0.5$; and \textbf{3) boundaries at assembly completion}, ending each subtask at capture. Fixed durations degrade all furniture, sharply IVAR, whose durations vary too much for a fixed budget. The binary head drops despite the matched switching point: look-alike frames around it carry opposite labels, so without a graded target it switches early or late, whereas continuous progress approaches $\tau_p$ smoothly. Completion boundaries drop and become threshold-sensitive, improving at $\tau_p{=}0.98$ where post-retreat boundaries do not: progress must cross $\tau_p$ within the few frames around capture, so $\tau_p{=}0.95$ switches before the part seats. Post-retreat boundaries give a long, visually distinct switching window.

\begin{table}[b]
\vspace{-0.5em}
\setlength{\abovecaptionskip}{0.1cm}
    \footnotesize
    \caption{\textbf{Sensitivity to alignment tolerance} (success rates~$\uparrow$). Translational tolerance is listed for small / large parts; rotational tolerance applies to all parts. Both are scaled by the same factor. Nominal FurnitureVLA setting in bold.}
    \begin{adjustbox}{width=1.0\linewidth,center}
    \centering
    \begin{tabular}{lcccccc}
    \toprule
    \multirow{2}{*}{\textbf{Scale}} & \multicolumn{2}{c}{\textbf{Tolerance}} & \multicolumn{4}{c}{\textbf{Success Rate}} \\
    \cmidrule(lr){2-3} \cmidrule(lr){4-7}
    & \textbf{Trans.\ (cm, small / large)} & \textbf{Rot.\ ($^\circ$)} & \textbf{LACK} & \textbf{KALLAX} & \textbf{IVAR} & \textbf{Average} \\ \noalign{\vskip 0.3ex}
    \hline \noalign{\vskip 1.0ex}
    $0.75\times$ & 0.75 / 1.5 & 3 & 0.94 & 0.79 & 0.52 & 0.75 \\
    \noalign{\vskip 0.5ex}
    $0.5\times$ & 0.5 / 1.0 & 2 & 0.80 & 0.68 & 0.46 & 0.65 \\
    \noalign{\vskip 0.5ex}
    $0.25\times$ & 0.25 / 0.5 & 1 & 0.27 & 0.18 & 0.09 & 0.18 \\
    \noalign{\vskip 0.3ex}
    \midrule \noalign{\vskip 0.2ex}
    \textbf{FurnitureVLA (1.0$\times$)} & \textbf{1.0 / 2.0} & \textbf{4} & \textbf{0.98} & \textbf{0.85} & \textbf{0.56} & \textbf{0.80} \\
    \bottomrule
    \end{tabular}
    \label{tab:tolerance}
\end{adjustbox}
\vspace{-0.5em}
\end{table}

\subsection{Design Factors to Improve Assembly 
Success}\label{sec:Design Factors to Improve Assembly Success}
\vspace{-0.25em}

We next ask what improves assembly success once coordination is handled. Perception and control-loop choices are known to affect generalist policies~\cite{aloha23, bai2025openpi}, and real-scale bimanual assembly tackles both: large, heavy parts amplify control instabilities, and bimanual interactions occlude the scene, so we isolate four such factors.

\vspace{0.5em}
\noindent\textbf{Setup.} \textbf{1) Action horizon:} $\pi_{0.5}$ predicts chunks of $H{=}50$ actions; we execute the first 5/10/25 before replanning, trading error recovery against overhead and trajectory smoothness. \textbf{2) Temporal ensembling:} following ACT~\cite{aloha23}, the executed action is a weighted average over a rolling buffer of $B$ overlapping chunks, with weight $w_i = e^{\lambda i}$ on the prediction made $i$ steps ago; more negative $\lambda$ favors recent predictions and $\lambda{=}0$ averages uniformly. The progress $\hat{p}_t$ uses the latest prediction. \textbf{3) Observation viewpoint:} removing the rear camera or replacing it with front-view depth. \textbf{4) Observation resolution:} $224{\times}224$, $300{\times}300$, or $448{\times}448$.

\vspace{0.5em}
\noindent\textbf{Impact of Design Factors.} Temporal ensembling consistently improves success, best at $\lambda{=}-0.1$ on all furniture (Table~\ref{tab:design_factors}), where $70\%$ of the weight falls on the most recent prediction: large, heavy parts and coordinated bimanual motion benefit from smoothing. Horizons of 10 and 25 perform best depending on the furniture; KALLAX favors 25, likely because its largest and heaviest parts need stronger smoothing. The rear camera outperforms front-view depth and no rear view, mitigating occlusion, and higher resolution improves success. See Appendix~\ref{sec:Additional Results} for per-subtask breakdown.

\begin{table}[t]
    \setlength{\abovecaptionskip}{0.1cm}
    \footnotesize
    \caption{\textbf{Failure breakdown} over 100 episodes per furniture (default setup). Each failed episode is assigned one physical cause; counts sum to the failed count. Transition errors (premature or missed subtask switches by the progress head) co-occur with these failures and are excluded from the sum.}
    \label{tab:failures}
    \centering
    \begin{adjustbox}{width=\linewidth,center}
    \renewcommand{\arraystretch}{1.15}
    \begin{tabular}{llcccc}
    \toprule
    \textbf{Category} & \textbf{Failure Mode}
        & \textbf{LACK} & \textbf{KALLAX}
        & \textbf{IVAR} & \textbf{Total} \\
    \midrule
    \multirow{4}{*}{\textbf{Physical}}
        & Grasp failure          & 1 & 6 & 28 & 35 \\
        & Toppled part           & 0 & 7 & 1  & 8  \\
        & Insertion misalignment & 1 & 2 & 13 & 16 \\
        & Transport drop         & 0 & 0 & 2  & 2  \\
    \midrule
    \multirow{2}{*}{\shortstack[l]{\textbf{Transition}\\\textbf{(co-occurring)}}}
        & Premature transition   & 0 & 5 & 13 & 18 \\
        & Missed transition      & 0 & 0  & 0  & 0  \\
    \midrule
    \multicolumn{2}{l}{\textbf{Failed / Evaluated}}
        & 2/100 & 15/100 & 44/100 & 61/300 \\
    \bottomrule
    \end{tabular}
    \end{adjustbox}
    \vspace{-1.0em}
\end{table}

\begin{table}[b]
\vspace{-0.5em}
\setlength{\abovecaptionskip}{0.1cm}
    \footnotesize
    \caption{\textbf{Effect of demonstration quantity} (success rates~$\uparrow$).}
    \begin{adjustbox}{width=1.0\linewidth,center}
    \centering
    \begin{tabular}{lcccc}
    \toprule
    \textbf{Method} & \textbf{LACK} & \textbf{KALLAX} & \textbf{IVAR} & \textbf{Average} \\ \noalign{\vskip 0.3ex}
    \hline \noalign{\vskip 1.0ex}
    FurnitureVLA (25$\%$ demos) & 0.53 & 0.63 & 0.33 & 0.50 \\ \noalign{\vskip 0.5ex}
    FurnitureVLA (50$\%$ demos) & 0.82 & 0.78 & 0.44 & 0.68 \\
    \noalign{\vskip 0.5ex}
    FurnitureVLA (100$\%$ demos) & \textbf{0.98} & \textbf{0.85} & \textbf{0.56} & \textbf{0.80} \\
    \bottomrule
    \end{tabular}
    \label{tab:ablations}
\end{adjustbox}
\vspace{-0.5em}
\end{table}

\begin{figure*}[t]
  \centering
  \footnotesize
  \setlength{\abovecaptionskip}{0.1cm}
  \includegraphics[width=1\linewidth]{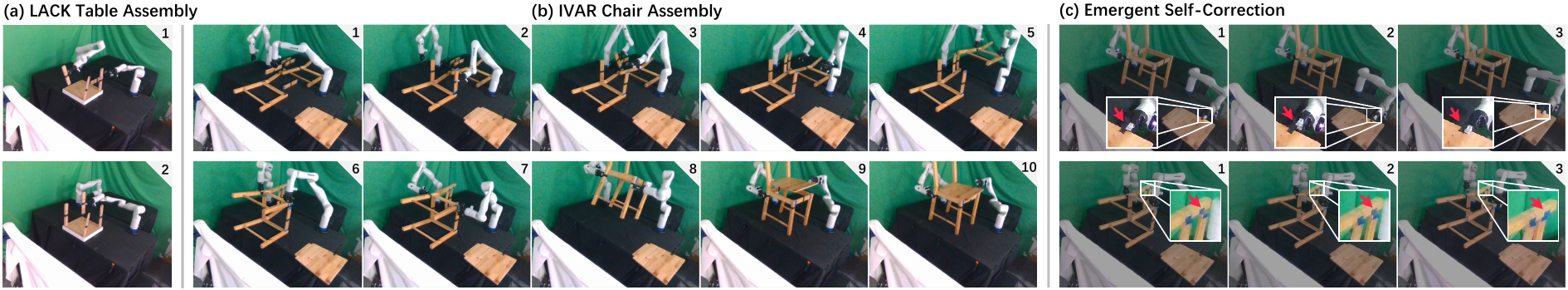}
  \vspace{-3mm} \caption{\textbf{Qualitative real-robot rollouts.} (a, b) Successful long-horizon assembly of the LACK side table and the IVAR chair. (c) Emergent self-correction behaviors: regrasping to improve contact and corrective motions to align the magnets.}
  \label{fig:demo}
  \vspace{-1.0em} 
\end{figure*}

\begin{figure}[t]
  \centering
  \footnotesize
  \setlength{\abovecaptionskip}{0.1cm}
  \includegraphics[width=1\linewidth]{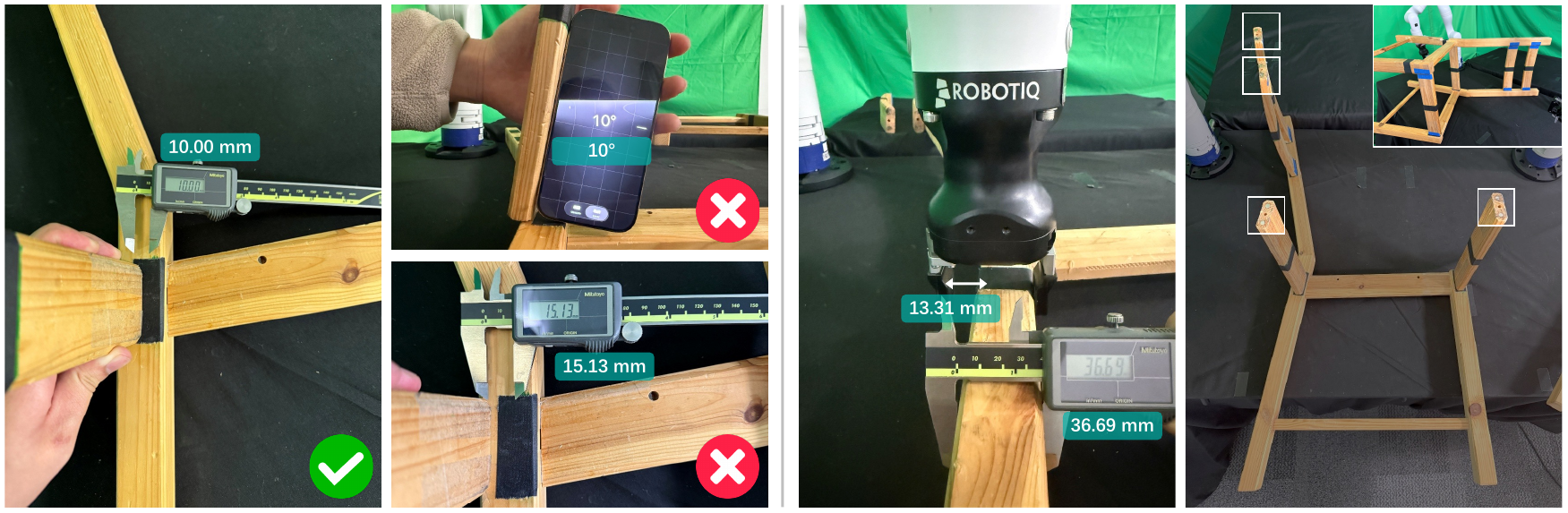}
  \vspace{-3mm} \caption{\textbf{Real-world alignment requirements.} (Left) assembly tolerance with magnets; successful attachment at $\sim$1\,cm offset. (Right) gripper clearance and multi-magnet alignment challenges.}
  \label{fig:eval}
  \vspace{-1.0em} 
\end{figure}

\subsection{What Bounds Assembly Success}\label{sec:What Bounds Assembly Success}
Even at the best design configuration, average success plateaus at 0.80. We ask what bounds the remainder from three sides: how small an alignment error the policy achieves, which failure modes dominate, and if more demonstrations help. Since the magnets fix the tolerance at the joint's capture range, nominal success shows only that the policy meets it; we therefore tighten the tolerance until the policy fails.

\vspace{0.5em}
\noindent\textbf{Sensitivity to Alignment Tolerance.} We scale the nominal translational and rotational tolerances (Sec.~\ref{sec:Bimanual Furniture Assembly Tasks}) by $0.75$, $0.5$ and $0.25$, tightening the capture condition and success check together (Table~\ref{tab:tolerance}). Success degrades gracefully to $0.5\times$ (0.80 to 0.65) and collapses at $0.25\times$ (0.18). Translation is the binding constraint: at capture, the insertion-axis error sits at roughly 90\% of its tolerance for most parts, whereas rotation stays well inside its bound (1--3$^\circ$ against 4$^\circ$).

\vspace{0.5em}
\noindent\textbf{Failure Breakdown.} Using privileged simulator state, we assign each failed rollout one physical cause from predicates (Appendix~\ref{sec:Analyses Details}) at its first uncompleted subtask (e.g., transport drop: a lifted part falls over 3\,cm), and separately record premature and missed transitions, i.e., switching before or not after completion (Table~\ref{tab:failures}). Grasp failures (35 of 61) and insertion misalignment (16) dominate, mostly on IVAR; toppling is specific to KALLAX's large, heavy parts. Premature transitions co-occur with missed grasps and toppling, which yield states unseen in training, yet remain limited, and no missed transitions occur. Completion recognition thus rarely fails outright, leaving physical execution, not switching, as the main limit. Appendix~\ref{sec:Analyses Details} validates the labels by human inspection and explores progress-triggered retry.

\vspace{0.5em}
\noindent\textbf{Effect of Demonstration Quantity.} Finally, we ask whether the bound is a data limit. Success improves consistently with more data (Table~\ref{tab:ablations}), with the largest gain from $25\%$ to $50\%$; the smaller gain from $50\%$ to $100\%$ suggests our demonstration budget balances collection cost and performance.

\section{Real-World Experiments}
\vspace{-0.25em}
We validate FurnitureVLA on the real bimanual Kinova Gen3 platform (Fig.~\ref{fig:system}) with the LACK side table and the hardest IVAR chair, asking: 1) whether long-horizon assembly transfers to real hardware; 2) whether real-world failures arise from individual subtasks or accumulate across them; and 3) whether co-training with simulation data helps.

\subsection{Full and Per-Part Assembly Performance}
\noindent\textbf{Setup and Evaluation.} We collect 50 demonstrations for LACK and 100 for IVAR via Quest2ROS~\cite{welle2024quest2ros} teleoperation, train as in simulation, and evaluate 15 rollouts each, with monolithic $\pi_{0.5}$ as the IVAR baseline. Magnets sit in the screw holes, but parts must be aligned: they snap together at a 1\,cm deviation yet separate at 1.5\,cm or a $10^\circ$ tilt (Fig.~\ref{fig:eval}, left), and limited gripper clearance (5\,cm Hand-E opening against part thickness up to 3.7\,cm) and engaging up to 8 magnets add constraints (Fig.~\ref{fig:eval}, right). We report full-assembly success as in Sec.~\ref{sec:Long-Horizon Assembly Performance}, and per-part success, where each subtask is evaluated independently from its intermediate state over 15 rollouts. Details in Appendix~\ref{sec:Real System Details}.

\begin{figure}[t]
  \centering
  \footnotesize
  \setlength{\abovecaptionskip}{0.1cm}
  \includegraphics[width=1\linewidth]{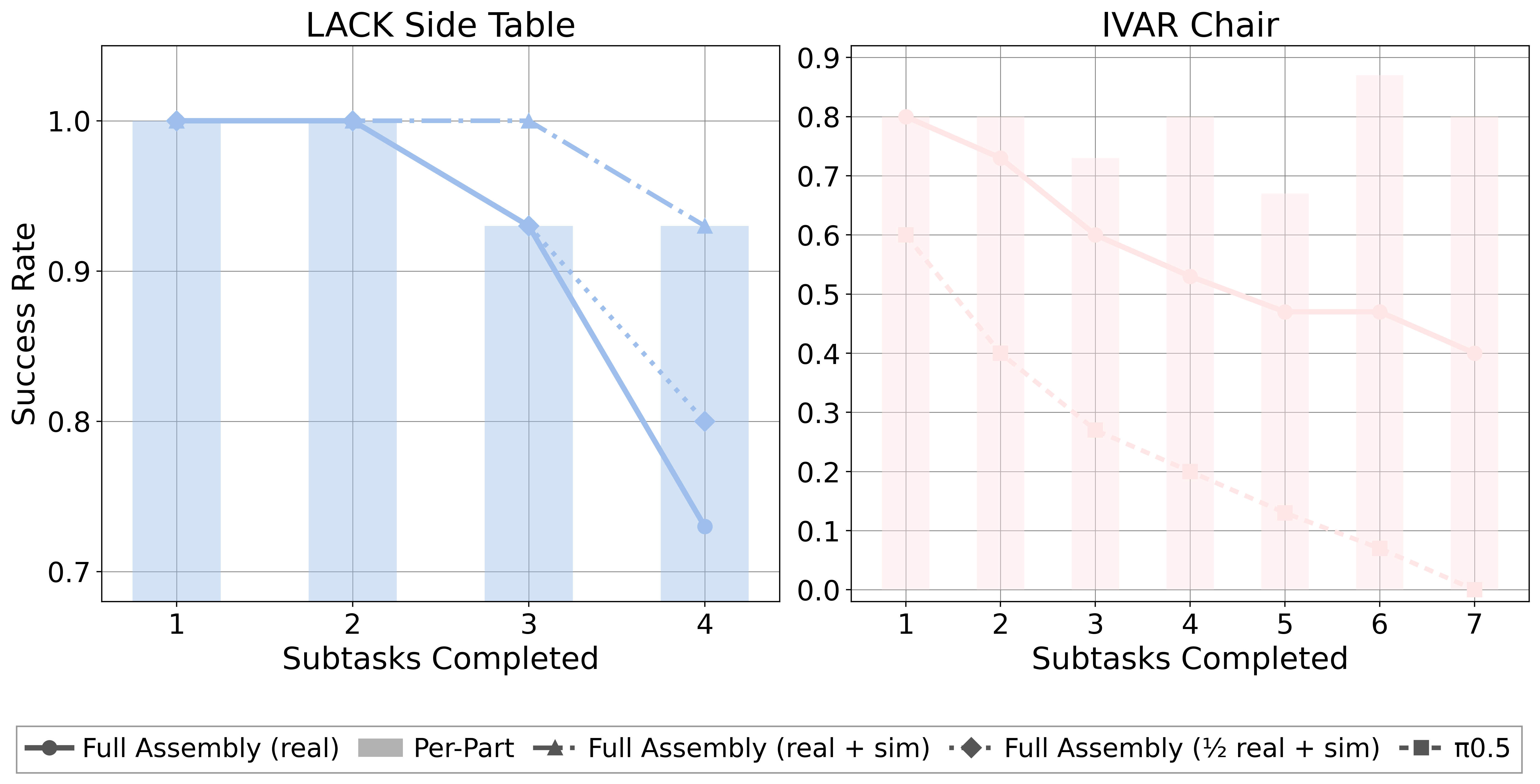}
  \vspace{-3mm} \caption{\textbf{Real-world assembly performance} (success rates~$\uparrow$). Lines show full-assembly success, which is monotonically decreasing across subtasks; bars show per-part success, where each subtask is evaluated independently from its intermediate state.}
  \label{fig:real_world}
  \vspace{-1.0em}
\end{figure}

\vspace{0.5em}
\noindent\textbf{Results.} Real-world full-assembly success is lower than in simulation (0.40 vs.\ 0.56 on IVAR, 0.73 vs.\ 0.98 on LACK; Fig.~\ref{fig:real_world} vs.\ Table~\ref{tab:long_horizon}) due to added difficulty and noise of real settings. Monolithic $\pi_{0.5}$ drops to 0 by the final IVAR subtask. IVAR subtasks 3--5 are the bottlenecks: 3 and 4 lose visual detail as the left arm operates farther from the camera, and 5 must bring several magnet pairs within tolerance; on LACK, failures are confined to the last two. Per-part success is higher than full-assembly success (on LACK, ${\geq}0.93$ against 0.73), so failures accumulate across subtasks rather than stemming from a single one.

\vspace{0.5em}
\noindent\textbf{Emergent Behavior and Qualitative Results.} Fig.~\ref{fig:demo} shows successful rollouts (a, b) and emergent self-correction (c): with insufficient contact on the seat panel, the robot reopens the gripper, adjusts its pose, and regrasps; when attaching the left side frame, it makes small corrective motions to align the magnets before insertion. We attribute these behaviors to similar corrections in the teleoperated demonstrations.


\subsection{Simulation and Real Data Co-Training}
\noindent\textbf{Setup and Evaluation.} Since our simulation and real-world setups are closely matched (Figs.~\ref{fig:tasks} and~\ref{fig:demo}), we test whether simulation data can substitute for real data on LACK. Against the real-only policy (50 real demonstrations), we co-train \textbf{Full} (50 real + 50 simulation) and \textbf{Half} (25 real + 50 simulation), each evaluated over 15 real-world rollouts.

\vspace{0.5em}
\noindent\textbf{Results.} Co-training on the full real set raises final full-assembly success from 0.73 to 0.93 (Fig.~\ref{fig:real_world}), with the gain concentrated in the later, contact-rich subtasks where the real-only policy degrades. Half still reaches 0.80, outperforming the real-only policy trained on twice as much real data. Simulation data thus complements real data and suggests it can partially substitute for it.

\section{Conclusion and Discussion}
We present \textbf{FurnitureVLA}, the first systematic study of real-scale bimanual furniture assembly with VLAs. We build a simulation and VR teleoperation system, propose a progress-enhanced VLA for improved long-horizon execution, and investigate what improves and what bounds assembly success in simulation and on a real robot.

\vspace{0.5em}
\noindent\textbf{Limitations.} Our fixed-base setup limits assembly to furniture within the robot's workspace, and magnets evaluate pre-insertion alignment rather than fastening; mobile bimanual platforms and tool use remain open challenges.

{
    \small
    \bibliographystyle{ieeenat_fullname}
    \bibliography{main}

@article{pi0_5_pi,
  author       = {{Physical Intelligence}},
  title        = {{\(\pi\)}\({}_{\mbox{0.5}}\): a Vision-Language-Action Model with
                  Open-World Generalization},
  journal      = {arXiv preprint arXiv:2504.16054},
  year         = {2025}
}

@inproceedings{FuZF24,
  author       = {Zipeng Fu and
                  Tony Z. Zhao and
                  Chelsea Finn},
  title        = {Mobile {ALOHA:} Learning Bimanual Mobile Manipulation using Low-Cost
                  Whole-Body Teleoperation},
  booktitle    = {Conference on Robot Learning},
  year         = {2024}
}

@article{li2025robotmover,
  title={RobotMover: Learning to Move Large Objects From Human Demonstrations},
  author={Li, Tianyu and Truong, Joanne and Yang, Jimmy and Clegg, Alexander and Rai, Akshara and Ha, Sehoon and Puig, Xavier},
  journal={arXiv preprint arXiv:2502.05271},
  year={2025}
}

@inproceedings{FurnitureBench23,
  author       = {Minho Heo and
                  Youngwoon Lee and
                  Doohyun Lee and
                  Joseph J. Lim},
  title        = {FurnitureBench: Reproducible Real-World Benchmark for Long-Horizon
                  Complex Manipulation},
  booktitle    = {Robotics: Science and Systems},
  year         = {2023}
}

@inproceedings{LinCZ24,
  author       = {Haohong Lin and
                  Radu Corcodel and
                  Ding Zhao},
  title        = {Generalize by Touching: Tactile Ensemble Skill Transfer for Robotic
                  Furniture Assembly},
  booktitle    = {International Conference on Robotics and Automation},
  year         = {2024}
}

@inproceedings{AnkileSS024,
  author       = {Lars Ankile and
                  Anthony Simeonov and
                  Idan Shenfeld and
                  Pulkit Agrawal},
  title        = {{JUICER:} Data-Efficient Imitation Learning for Robotic Assembly},
  booktitle    = {International Conference on Intelligent Robots and Systems},
  year         = {2024}
}

@inproceedings{tian2025fabrica,
  title={Fabrica: Dual-Arm Assembly of General Multi-Part Objects via Integrated Planning and Learning},
  author={Tian, Yunsheng and Jacob, Joshua and Huang, Yijiang and Zhao, Jialiang and Gu, Edward and Ma, Pingchuan and Zhang, Annan and Javid, Farhad and Romero, Branden and Chitta, Sachin and others},
  booktitle={Conference on Robot Learning},
  year={2025}
}

@inproceedings{ZhangWSWZT23,
  author       = {Xiang Zhang and
                  Changhao Wang and
                  Lingfeng Sun and
                  Zheng Wu and
                  Xinghao Zhu and
                  Masayoshi Tomizuka},
  title        = {Efficient Sim-to-real Transfer of Contact-Rich Manipulation Skills
                  with Online Admittance Residual Learning},
  booktitle    = {Conference on Robot Learning},
  year         = {2023}
}

@inproceedings{AnkileSSTA25,
  author       = {Lars Ankile and
                  Anthony Simeonov and
                  Idan Shenfeld and
                  Marcel Torne and
                  Pulkit Agrawal},
  title        = {From Imitation to Refinement - Residual Rl for Precise Assembly},
  booktitle    = {International Conference on Robotics and Automation},
  year         = {2025}
}

@article{NoseworthyTWHKRFRNA25,
  author       = {Michael Noseworthy and
                  Bingjie Tang and
                  Bowen Wen and
                  Ankur Handa and
                  Chad C. Kessens and
                  Nicholas Roy and
                  Dieter Fox and
                  Fabio Ramos and
                  Yashraj Narang and
                  Iretiayo Akinola},
  title        = {{FORGE:} Force-Guided Exploration for Robust Contact-Rich Manipulation
                  Under Uncertainty},
  journal      = {Robotics and Automation Letters},
  year         = {2025}
}

@inproceedings{VTRefine_huang,
  author       = {Binghao Huang and
                  Jie Xu and
                  Iretiayo Akinola and
                  Wei Yang and
                  Balakumar Sundaralingam and
                  Rowland O'Flaherty and
                  Dieter Fox and
                  Xiaolong Wang and
                  Arsalan Mousavian and
                  Yu{-}Wei Chao and
                  Yunzhu Li},
  title        = {VT-Refine: Learning Bimanual Assembly with Visuo-Tactile Feedback
                  via Simulation Fine-Tuning},
  booktitle    = {Conference on Robot Learning},
  year         = {2025}
}

@inproceedings{KimPKXB0RFSVKBT24,
  author       = {Moo Jin Kim and
                  Karl Pertsch and
                  Siddharth Karamcheti and
                  Ted Xiao and
                  Ashwin Balakrishna and
                  Suraj Nair and
                  Rafael Rafailov and
                  Ethan Paul Foster and
                  Pannag R. Sanketi and
                  Quan Vuong and
                  Thomas Kollar and
                  Benjamin Burchfiel and
                  Russ Tedrake and
                  Dorsa Sadigh and
                  Sergey Levine and
                  Percy Liang and
                  Chelsea Finn},
  title        = {OpenVLA: An Open-Source Vision-Language-Action Model},
  booktitle    = {Conference on Robot Learning},
  year         = {2024},
}

@misc{generalist2025gen0,
  author       = {{Generalist AI Team}},
  title        = {GEN-0: Embodied Foundation Models That Scale with Physical Interaction},
  year         = {2025},
  howpublished = {\url{https://generalistai.com/blog/nov-04-2025-GEN-0}}
}

@inproceedings{aloha23,
  author       = {Tony Z. Zhao and
                  Vikash Kumar and
                  Sergey Levine and
                  Chelsea Finn},
  title        = {Learning Fine-Grained Bimanual Manipulation with Low-Cost Hardware},
  booktitle    = {Robotics: Science and Systems},
  year         = {2023}
}

@inproceedings{RossGB11,
  author       = {St{\'{e}}phane Ross and
                  Geoffrey J. Gordon and
                  Drew Bagnell},
  title        = {A Reduction of Imitation Learning and Structured Prediction to No-Regret
                  Online Learning},
  booktitle    = {International Conference on Artificial
                  Intelligence and Statistics},
  year         = {2011}
}

@inproceedings{sun2024arch,
  title={ARCH: Hierarchical hybrid learning for long-horizon contact-rich robotic assembly},
  author={Sun, Jiankai and Curtis, Aidan and You, Yang and Xu, Yan and Koehle, Michael and Chen, Qianzhong and Huang, Suning and Guibas, Leonidas and Chitta, Sachin and Schwager, Mac and others},
  booktitle    = {Conference on Robot Learning},
  year={2025}
}

@inproceedings{ShiIEKPVTWWFLDG25,
  author       = {Lucy Xiaoyang Shi and
                  Brian Ichter and
                  Michael Robert Equi and
                  Liyiming Ke and
                  Karl Pertsch and
                  Quan Vuong and
                  James Tanner and
                  Anna Walling and
                  Haohuan Wang and
                  Niccolo Fusai and
                  Adrian Li{-}Bell and
                  Danny Driess and
                  Lachy Groom and
                  Sergey Levine and
                  Chelsea Finn},
  title        = {Hi Robot: Open-Ended Instruction Following with Hierarchical Vision-Language-Action
                  Models},
  booktitle    = {International Conference on Machine Learning},
  year         = {2025}
}

@article{ZhouP18,
  author       = {Xian Zhou and
                  Quang{-}Cuong Pham},
  title        = {Can robots assemble an {IKEA} chair?},
  journal      = {Science Robotics},
  year         = {2018}
}

@inproceedings{tie2025manual,
  title     = {Manual2Skill: Learning to Read Manuals and Acquire Robotic Skills for Furniture Assembly Using Vision-Language Models},
  author    = {Tie, Chenrui and Sun, Shengxiang and Zhu, Jinxuan and Liu, Yiwei and Guo, Jingxiang and Hu, Yue and Chen, Haonan and Chen, Junting and Wu, Ruihai and Shao, Lin},
  booktitle   = {Robotics: Science and Systems},
  year      = {2025}
}

@inproceedings{LeeHL21,
  author       = {Youngwoon Lee and
                  Edward S. Hu and
                  Joseph J. Lim},
  title        = {{IKEA} Furniture Assembly Environment for Long-Horizon Complex Manipulation
                  Tasks},
  booktitle    = {International Conference on Robotics and Automation},
  year         = {2021}
}

@inproceedings{TangLAHS0FN23,
  author       = {Bingjie Tang and
                  Michael A. Lin and
                  Iretiayo Akinola and
                  Ankur Handa and
                  Gaurav S. Sukhatme and
                  Fabio Ramos and
                  Dieter Fox and
                  Yashraj Narang},
  title        = {IndustReal: Transferring Contact-Rich Assembly Tasks from Simulation
                  to Reality},
  booktitle    = {Robotics: Science and Systems},
  year         = {2023}
}

@inproceedings{GiacomuzzoTJR24,
  author       = {Giulio Giacomuzzo and
                  Matteo Terreran and
                  Siddarth Jain and
                  Diego Romeres},
  title        = {{DECAF:} a Discrete-Event based Collaborative Human-Robot Framework
                  for Furniture Assembly},
  booktitle    = {International Conference on Intelligent Robots and Systems},
  year         = {2024}
}

@article{ma2026cyclevla,
  title={CycleVLA: Proactive Self-Correcting Vision-Language-Action Models via Subtask Backtracking and Minimum Bayes Risk Decoding},
  author={Ma, Chenyang and Yang, Guangyu and Lu, Kai and Xu, Shitong and Byrne, Bill and Trigoni, Niki and Markham, Andrew},
  journal={arXiv preprint arXiv:2601.02295},
  year={2026}
}

@inproceedings{lu2025kitchenvla,
  author    = {K. Lu and C. Ma and C. Hori and D. Romeres},
  title     = {{KitchenVLA}: Iterative Vision-Language Corrections for Robotic Execution of Human Tasks},
  booktitle = {International Conference on Robotics and Automation Workshop Safe-VLM},
  year      = {2025}
}

@inproceedings{ma2025coopera,
  title     = {COOPERA: Continual Open-Ended Human-Robot Assistance},
  author    = {Ma, Chenyang and Lu, Kai and Desai, Ruta and Puig, Xavier and Markham, Andrew and Trigoni, Niki},
  booktitle = {Advances in Neural Information Processing Systems},
  year      = {2025},
}

@article{YangZDBS25,
  author       = {Yue Yang and
                  Linfeng Zhao and
                  Mingyu Ding and
                  Gedas Bertasius and
                  Daniel Szafir},
  title        = {{BOSS:} Benchmark for Observation Space Shift in Long-Horizon Task},
  journal      = {Robotics and Automation Letters},
  year         = {2025}
}

@inproceedings{fan2025longvla,
      title={Long-VLA: Unleashing Long-Horizon Capability of Vision Language Action Model for Robot Manipulation}, 
      author={Yiguo Fan and Pengxiang Ding and Shuanghao Bai and Xinyang Tong and Yuyang Zhu and Hongchao Lu and Fengqi Dai and Wei Zhao and Yang Liu and Siteng Huang and Zhaoxin Fan and Badong Chen and Donglin Wang},
      year={2025},
      booktitle    = {Conference on Robot Learning},
}

@inproceedings{ZawalskiCPMFL24,
  author       = {Michal Zawalski and
                  William Chen and
                  Karl Pertsch and
                  Oier Mees and
                  Chelsea Finn and
                  Sergey Levine},
  title        = {Robotic Control via Embodied Chain-of-Thought Reasoning},
  booktitle    = {Conference on Robot Learning},
  year         = {2024}
}

@article{yang2026lilo,
  title={LiLo-VLA: Compositional Long-Horizon Manipulation via Linked Object-Centric Policies},
  author={Yang, Yue and Cheng, Shuo and Fang, Yu and Bharadhwaj, Homanga and Ding, Mingyu and Bertasius, Gedas and Szafir, Daniel},
  journal={arXiv preprint arXiv:2602.21531},
  year={2026}
}

@article{bai2025openpi,
  title={Openpi Comet: Competition Solution For 2025 BEHAVIOR Challenge},
  author={Bai, Junjie and Chao, Yu-Wei and Chen, Qizhi and Gu, Jinwei and Kim, Moo Jin and Li, Zhaoshuo and Li, Xuan and Lin, Tsung-Yi and Liu, Ming-Yu and Ma, Nic and others},
  journal={arXiv preprint arXiv:2512.10071},
  year={2025}
}

@inproceedings{ONeillRMGPLPGMJ24,
  author       = {{Open X-Embodiment Collaboration}},
  title        = {Open X-Embodiment: Robotic Learning Datasets and {RT-X} Models : Open
                  X-Embodiment Collaboration},
  booktitle    = {International Conference on Robotics and Automation},
  year         = {2024}
}

@article{pertsch2025fast,
  title={Fast: Efficient action tokenization for vision-language-action models},
  author={{Physical Intelligence}},
  journal={arXiv preprint arXiv:2501.09747},
  year={2025}
}

@inproceedings{atreya2025roboarena,
  title = {RoboArena: Distributed Real-World Evaluation of Generalist Robot Policies},
  author = {Atreya, Pranav and Pertsch, Karl and Lee, Tony and Kim, Moo Jin and Jain, Arhan and Kuramshin, Artur and Eppner, Clemens and Neary, Cyrus and Hu, Edward and Ramos, Fabio and others},
  booktitle = {Conference on Robot Learning},
  year = {2025}
}

@inproceedings{ma2024spatialpin,
  title={SpatialPIN: Enhancing Spatial Reasoning Capabilities of Vision-Language Models through Prompting and Interacting 3D Priors},
  author={Ma, Chenyang and Lu, Kai and Cheng, Ta-Ying and Trigoni, Niki and Markham, Andrew},
  booktitle={Neural Information Processing Systems},
  year={2024}
}

@article{seqvla_yang,
  author       = {Ran Yang and
                  Zijian An and
                  Lifeng Zhou and
                  Yiming Feng},
  title        = {SeqVLA: Sequential Task Execution for Long-Horizon Manipulation with
                  Completion-Aware Vision-Language-Action Model},
  journal      = {arXiv preprint arXiv:2509.14138},
  year         = {2025}
}

@article{gr00t_fan,
  author       = {{GR00T Team}},
  title        = {{GR00T} {N1:} An Open Foundation Model for Generalist Humanoid Robots},
  journal      = {arXiv preprint arXiv:2503.14734},
  year         = {2025}
}

@inproceedings{TangJ23,
  author       = {Ling Tang and
                  Yan{-}Bin Jia},
  title        = {Robotic Fastening with a Manual Screwdriver},
  booktitle    = {International Conference on Robotics and Automation},
  year         = {2023}
}

@article{HanCSJLCPB22,
  author       = {Sangchul Han and
                  Myoung{-}Su Choi and
                  Yong{-}Woo Shin and
                  Ga{-}Ram Jang and
                  Dong{-}Hyuk Lee and
                  Jungsan Cho and
                  Jae{-}Han Park and
                  Ji{-}Hun Bae},
  title        = {Screwdriving Gripper That Mimics Human Two-Handed Assembly Tasks},
  journal      = {Robotics},
  year         = {2022}
}

@inproceedings{LiuZGFLZS23,
  author       = {Bo Liu and
                  Yifeng Zhu and
                  Chongkai Gao and
                  Yihao Feng and
                  Qiang Liu and
                  Yuke Zhu and
                  Peter Stone},
  title        = {{LIBERO:} Benchmarking Knowledge Transfer for Lifelong Robot Learning},
  booktitle    = {Advances in Neural Information Processing Systems},
  year         = {2023}
}

@article{MeesHRB22,
  author       = {Oier Mees and
                  Luk{\'{a}}s Hermann and
                  Erick Rosete{-}Beas and
                  Wolfram Burgard},
  title        = {{CALVIN:} {A} Benchmark for Language-Conditioned Policy Learning for
                  Long-Horizon Robot Manipulation Tasks},
  journal      = {Robotics and Automation Letters},
  year         = {2022}
}

@inproceedings{robocasa2024,
  title={RoboCasa: Large-Scale Simulation of Everyday Tasks for Generalist Robots},
  author={Soroush Nasiriany and Abhiram Maddukuri and Lance Zhang and Adeet Parikh and Aaron Lo and Abhishek Joshi and Ajay Mandlekar and Yuke Zhu},
  booktitle={Robotics: Science and Systems},
  year={2024}
}

@misc{3dwarehouse,
  author       = {{3D Warehouse}},
  title        = {3D Warehouse},
  howpublished = {\url{https://3dwarehouse.sketchup.com/}},
  year         = {2026}
}

@misc{ambientcg,
  author       = {{ambientCG}},
  title        = {ambientCG},
  howpublished = {\url{https://ambientcg.com/}},
  year         = {2026}
}

@inproceedings{MakoviychukWGLS21,
  author       = {Viktor Makoviychuk and
                  Lukasz Wawrzyniak and
                  Yunrong Guo and
                  Michelle Lu and
                  Kier Storey and
                  Miles Macklin and
                  David Hoeller and
                  Nikita Rudin and
                  Arthur Allshire and
                  Ankur Handa and
                  Gavriel State},
  title        = {Isaac Gym: High Performance {GPU} Based Physics Simulation For Robot
                  Learning},
  booktitle    = {NeurIPS Datasets and Benchmarks},
  year         = {2021}
}

@misc{blender_2025,
  author = {{Blender Foundation}},
  title = {Blender},
  howpublished = {\url{https://www.blender.org}},
  year = {2025}
}

@inproceedings{welle2024quest2ros,
title={Quest2ROS: An App to Facilitate Teleoperating Robots},
author={Welle, Michael C and Ingelhag, Nils and Lippi, Martina and Wozniak, Maciej and Gasparri, Andrea and Kragic, Danica},
booktitle={Workshop on VAM-HRI},
year={2024}}

@inproceedings{KhazatskyP0BDKN24,
  author       = {{DROID Team}},
  title        = {{DROID:} {A} Large-Scale In-The-Wild Robot Manipulation Dataset},
  booktitle    = {Robotics: Science and Systems},
  year         = {2024}
}

@inproceedings{chi2023diffusion,
  title     = {Diffusion Policy: Visuomotor Policy Learning via Action Diffusion},
  author    = {Chi, Cheng and Xu, Zhenjia and Feng, Siyuan and Cousineau, Eric and Du, Yilun and Burchfiel, Benjamin and Tedrake, Russ and Song, Shuran},
  booktitle = {Robotics: Science and Systems (RSS)},
  year      = {2023}
}

@misc{GPT-6-astra_openai,
  title={GPT-6 Astra: A new generation of intelligence
},
  author = {Open AI},
  howpublished={\url{https://openai.com/index/gpt-6-astra}},
  year = {2026}
}

@article{lu2026pgx,
  title={PGX: Physically Grounded Execution for Long-Horizon Humanoid Loco-Manipulation},
  author={Lu, Kai and Ma, Chenyang and Havoutis, Ioannis},
  journal={arXiv preprint arXiv:2611.26596},
  year={2026}
}
}

\ifsupp
\clearpage
\maketitlesupplementary

\setcounter{section}{0}
\renewcommand{\thesection}{\Alph{section}}
\renewcommand{\thesubsection}{\thesection.\arabic{subsection}}
\setcounter{secnumdepth}{2}
\renewcommand{\theHsection}{appendix.\Alph{section}}

\section{Overview}
This appendix includes: 1) additional details on constructing the simulation environment, 2) details of the real-world system, 3) additional details of FurnitureVLA, 4) implementation details of the baselines, and 5) additional analyses, together with a breakdown of the simulation results.

\begin{figure}[t]
  \centering
  \footnotesize
  \setlength{\abovecaptionskip}{0.1cm}
  \includegraphics[width=1\linewidth]{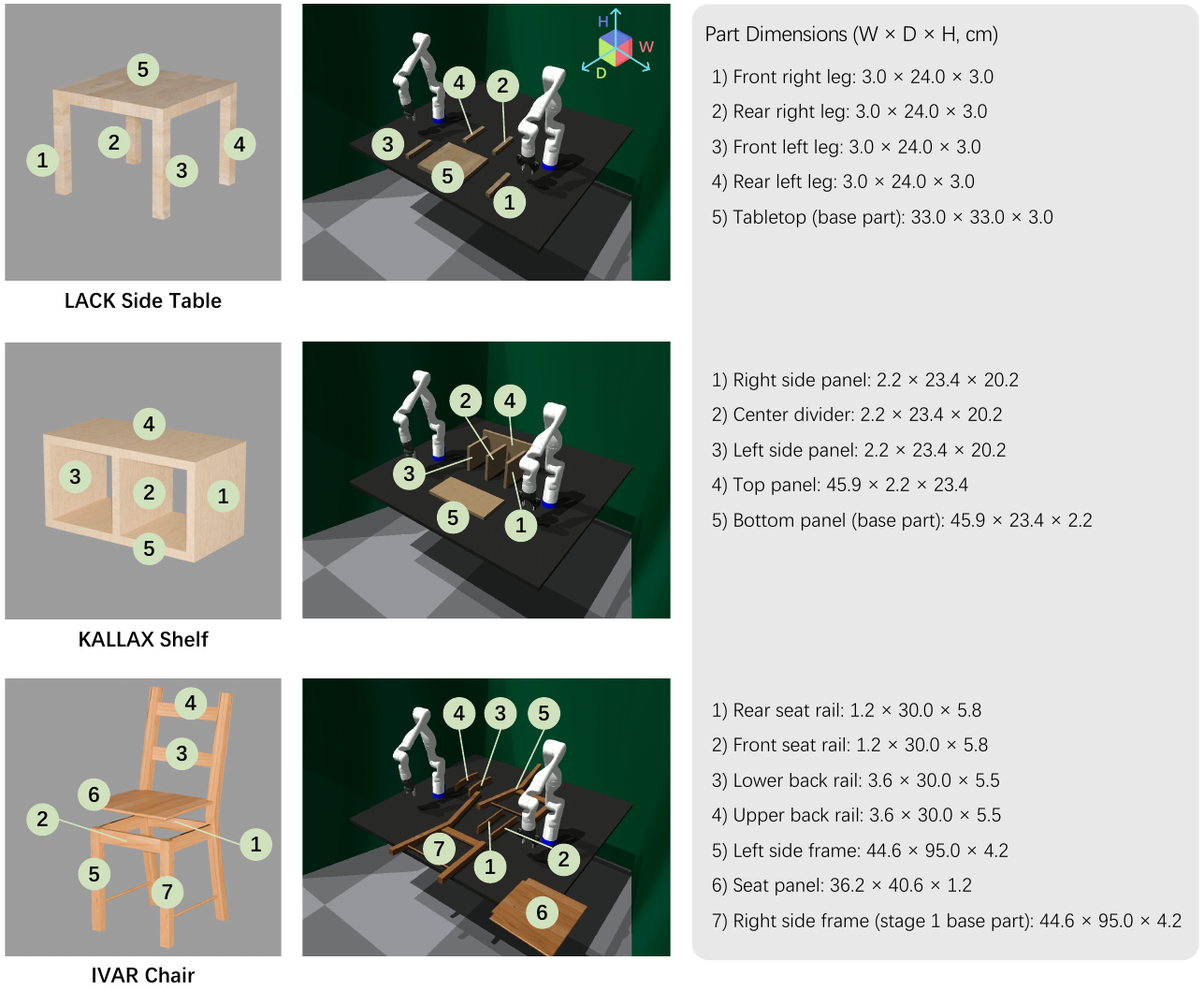}
  \vspace{-3mm} \caption{\textbf{Furniture part taxonomy and geometric properties.} Illustration of part naming, dimensions, and base part definition used in assembly. For the IVAR chair stage~1 definition, see Sec.~\ref{sec:Simulation Implementation Details}.}
  \label{fig:furniture_models}
  \vspace{-1.0em}
\end{figure}

\section{Additional Details of Simulation}\label{sec:Additional Details of Simulation}

\subsection{Furniture Models}
The IKEA furniture used can be purchased online. 3D models and textures are also publicly available and can be used for reproducibility. IKEA serial IDs and links are below:

\vspace{0.5em}
\begin{itemize}
\renewcommand\labelitemi{\scalebox{0.75}{$\bullet$}}
\item \textbf{LACK side table}: [\href{https://www.ikea.com/gb/en/p/lack-side-table-white-30514791/}{\textcolor{magenta}{30514791}}] [\href{https://3dwarehouse.sketchup.com/model/41b9b8677a6b5d07f51fa0238791f5dc/IKEA-LACK-Side-table-white}{\textcolor{magenta}{3D model}}] [\href{https://ambientcg.com/view?id=Wood083A}{\textcolor{magenta}{textures}}]
\item \textbf{KALLAX shelf}:[\href{https://www.ikea.com/gb/en/p/kallax-shelving-unit-white-90301555/}{\textcolor{magenta}{90301555}}] [\href{https://3dwarehouse.sketchup.com/model/f2321f2f-7fa2-44e1-94ac-4db122f92b0f/KALLAX-2x1}{\textcolor{magenta}{3D model}}] [\href{https://ambientcg.com/view?id=Wood021}{\textcolor{magenta}{textures}}]
\item \textbf{IVAR chair}: [\href{https://www.ikea.com.tr/en/product/ivar-pine-wooden-chair-90263902/}{\textcolor{magenta}{90263902}}] [\href{https://3dwarehouse.sketchup.com/model/b928b5b8-c283-4ea6-9b29-02f0d7d2686a/IVAR-Chair-2-Electric-boogaloo}{\textcolor{magenta}{3D model}}] [\href{https://ambientcg.com/view?id=Wood092}{\textcolor{magenta}{textures}}]
\end{itemize}
\vspace{0.5em}

We illustrate the naming of each furniture part, their dimensions, and the definition of the base part in Fig.~\ref{fig:furniture_models}. Part names are manually defined based on their spatial location relative to the base part in the assembled configuration, rather than their positions on the tabletop. In simulation, all parts are assigned a density of pine wood ($450\,\text{kg/m}^3$), consistent with real IKEA components. Note that the structure and dimensions given by the linked furniture IDs are the official specifications; the simulator models are post-processed and adjusted to match them. For the KALLAX shelf, since the IKEA product was out of stock at the time of purchase, we instead use a functionally equivalent but smaller two-compartment shelf in the real world (\href{https://online-moebel-kaufen.de/shop/Baby-Kinderzimmer/Regale/Filou-LE25-von-POL-Power-Haengebox-Regal-weiss-mit-2-Faechern::102937.html}{POL Power Filou-LE25, 35-666}), and scale the simulator model accordingly. More details follow in Sec.~\ref{sec:Simulation Implementation Details}.

\subsection{Simulation Implementation Details}\label{sec:Simulation Implementation Details}
\noindent\textbf{Simulator and Assets.} We use Isaac Gym~\cite{MakoviychukWGLS21},
extending the FurnitureBench~\cite{FurnitureBench23} codebase, with furniture 3D models from 3D Warehouse~\cite{3dwarehouse}, textures from ambientCG~\cite{ambientcg}, and assets post-processed in
Blender 5.0~\cite{blender_2025}.

\begin{figure}[t]
  \centering
  \footnotesize
  \setlength{\abovecaptionskip}{0.1cm}
  \includegraphics[width=1\linewidth]{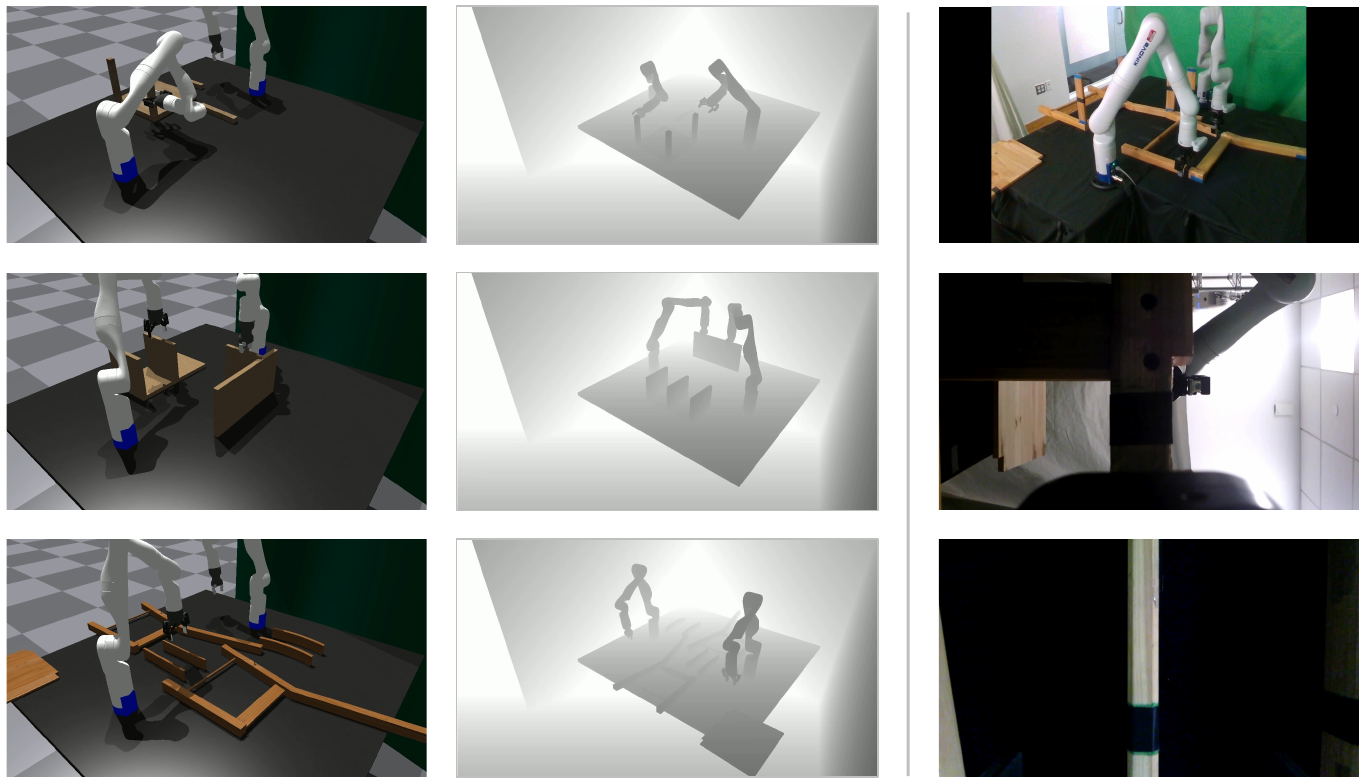}
  \vspace{-3mm} \caption{\textbf{Simulation and real-world observation examples.} (Left)~simulation rear-view RGB and depth observations. (Right)~real-world observations, including a rear-view image captured by an Intel RealSense D435 and two wrist-mounted images from the Kinova arms. All images are raw and unprocessed.}
  \label{fig:obv_example}
  \vspace{-1.0em}
\end{figure}

\vspace{0.5em}
\noindent\textbf{Simulation of Magnets.} Because Isaac Gym does not support runtime weld constraints, we simulate magnetic attachment by pose-resetting: at every \texttt{env.step()}, once the relative pose between a moving part and its target falls within $\epsilon$ translational and $\delta$ rotational tolerance, the moving part is kinematically teleported to maintain its rigid relative pose to the base part
on every subsequent tick, with linear and angular velocities synchronized.

\vspace{0.5em}
\noindent\textbf{IVAR Chair Two-Stage Simulation.} Isaac Gym does not support runtime weld constraints, so once the left side frame, rails, and right side frame are held together by the pose-reset magnets, the simulator cannot treat the resulting sub-structure as a single rigid body, preventing the final rotation required to make the partially assembled chair frame upright and attach the seat panel. We therefore split the IVAR chair episode into two stages. Stage~1 uses the right side frame as the base part and assembles the rails and left side frame against it. Stage~2 inherits the recorded terminal poses of the seat panel and partially assembled chair frame from Stage~1, then respawns the partially assembled chair frame as a single rigid mesh. In stage 2, the world frame is treated as the base reference. This allows us to evaluate whether lifting, rotating, and lowering the chair (subtask 6 in Fig.~\ref{fig:tasks}) is successful based on the relative pose error with respect to the world frame. Conceptually the task remains a single continuous assembly; the staging is purely an engineering workaround for the absence of weld constraints.

\vspace{0.5em}
\noindent\textbf{Control and Execution.}
The dual-arm Kinova Gen3 is controlled via an operational-space PD controller in the end-effector frame at $10\,\text{Hz}$.

\begin{figure}[t]
  \centering
  \footnotesize
  \setlength{\abovecaptionskip}{0.1cm}
  \includegraphics[width=1\linewidth]{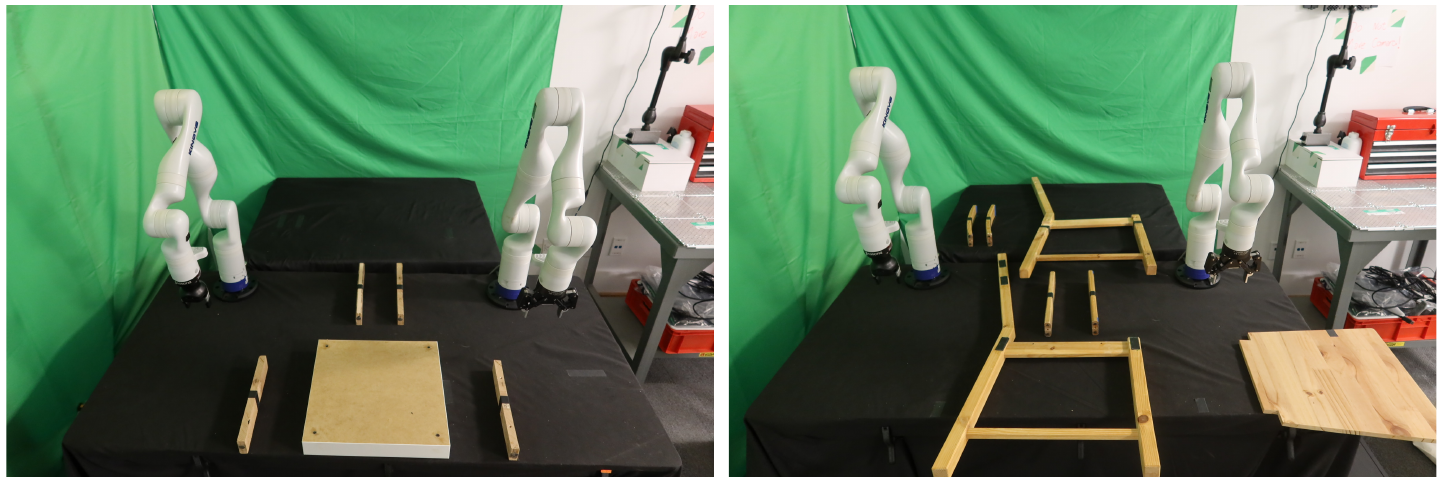}
  \vspace{-3mm} \caption{\textbf{LACK side table and IVAR chair assembly initializations.}}
  \label{fig:ivar_init}
  \vspace{-1.0em}
\end{figure}

\vspace{0.5em}
\noindent\textbf{Observations.}
Each environment provides RGB-D observations from four cameras: wrist-mounted cameras rigidly attached to the left and right end-effectors, along with a front and a rear camera. All observations are captured at a native resolution of $1280\times720$. Depth is measured using clipping planes at $0.001\,\text{m}$ and $3.5\,\text{m}$; values are clipped to $[0.001, 3.5]\,\text{m}$ and linearly normalized to the $[0,255]$ range. Fig.~\ref{fig:obv_example} shows examples of rear-view RGB and depth observations in simulation.

\vspace{0.5em}
\noindent\textbf{Data Generation Thresholds and Budgets.} Scripted demonstrations are generated via per-part motion planning. The relative pose threshold for successful assembly is set to $\epsilon = 5\,\text{mm}$ for translation and $\delta = 0.999$ (${\approx}3.0^\circ$ per rotation axis) for rotation. Each episode is capped at a furniture-specific step budget: 900 steps for IVAR chair stage~1, 650 for IVAR chair stage~2, 650 for the LACK side table, and 850 for the KALLAX shelf. Episodes are terminated and discarded if not all parts are assembled within this budget.

\vspace{0.5em}
\noindent\textbf{VLA Policy Evaluation.} As detailed in Sec.~\ref{sec:Simulation Data Generation and Success Criteria}, for VLA policy evaluation we use more relaxed thresholds than for data generation: $\epsilon = 1\,\text{cm}$ for small parts and $\epsilon = 2\,\text{cm}$ for large parts, with $\delta = 0.998$ (${\approx}4.0^\circ$ per rotation axis). The part classification is as follows:

\vspace{0.5em}
\begin{itemize}
\renewcommand\labelitemi{\scalebox{0.75}{$\bullet$}}
\item \textbf{Small parts:} LACK (front right leg, rear right leg, front left leg, rear left leg); KALLAX (right side panel, center divider, left side panel); IVAR (rear seat rail, front seat rail, lower back rail, upper back rail)
\item \textbf{Large parts:} KALLAX (top panel); IVAR (left side frame, partially assembled chair frame, seat panel)
\end{itemize}
\vspace{0.5em}

For evaluation, we extend the data generation budgets by 100 steps to 1000 (IVAR chair stage~1), 750 (IVAR chair stage~2), 750 (LACK side table), and 950 (KALLAX shelf).

\begin{table*}[t]
\setlength{\abovecaptionskip}{0.1cm}
\footnotesize
\caption{\textbf{Complete set of skill primitives.} Every subtask of every furniture item is a sequence of these primitives, implemented as states of the scripted per-part planners; the per-subtask primitive counts are the ``skills'' reported in Fig.~\ref{fig:tasks} (LACK 12, KALLAX 14, IVAR 25).}
\begin{adjustbox}{width=1.0\linewidth,center}
\centering
\begin{tabular}{@{}p{3.4cm}p{7.6cm}p{5.4cm}@{}}
\toprule
\textbf{Primitive} & \textbf{Description} & \textbf{Used by} \\
\noalign{\vskip 0.3ex}
\hline
\noalign{\vskip 1.0ex}
pick (single arm) & reach above the part, match the grasp orientation, descend, close the gripper, lift & LACK legs (4), KALLAX side panels and center divider (3), IVAR rails (4) \\
\noalign{\vskip 0.5ex}
place / insert (single arm) & move above the target, match the insertion orientation, lift to the pre-insertion height, descend to the hole or slot, insert, open the gripper & same parts \\
\noalign{\vskip 0.5ex}
retreat (single arm) & open the gripper and move up and back to a contact-free pose & every non-final subtask \\
\noalign{\vskip 0.5ex}
bimanual grasp & both arms reach above the part, align orientation, descend, and close (with right-arm avoidance of the assembled structure for IVAR) & KALLAX top panel, IVAR left side frame \\
\noalign{\vskip 0.5ex}
bimanual lift-and-attach & lift the held part, translate to the assembly position, rotate to the assembly orientation, lower, insert, release, retreat & KALLAX top panel (subtask~4), IVAR left side frame (subtask~5) \\
\noalign{\vskip 0.5ex}
bimanual lift, rotate upright, and place & approach, grasp with both arms, lift, rotate the sub-assembly by $90^\circ$ about a horizontal axis, align yaw, lower, move to the seat location & IVAR partially assembled chair frame (subtask~6) \\
\noalign{\vskip 0.5ex}
side grasp and horizontal attach (single arm) & approach from the side, rotate the wrist to a horizontal grasp, grasp, lift, translate, align, and push the part onto the frame & IVAR seat panel (subtask~7) \\
\bottomrule
\end{tabular}
\label{tab:skills}
\end{adjustbox}
\vspace{-0.5em}
\end{table*}

\vspace{0.5em}
\noindent\textbf{Complete Set of Skill Types.}
Table~\ref{tab:skills} lists every manipulation primitive used by the scripted planners, and therefore every skill the demonstrations contain. Single-arm insertion subtasks consist of pick, place, and retreat (3 skills); the bimanual subtasks compose grasp, lift, align-and-attach, and retreat (4--5 skills); and the IVAR frame flip composes six. These counts yield the 12, 14, and 25 skills per assembly reported in Fig.~\ref{fig:tasks}.

\section{Additional Details of Real System}\label{sec:Real System Details}


\noindent\textbf{Control and Execution.}
We operate both arms in end-effector space. At each 10\,Hz control step, the system computes a target end-effector pose delta for each arm, converts it into a Cartesian twist, and executes it via the Kinova Cartesian velocity controller, which performs low-level joint-velocity tracking on-board. Data collection and VLA inference share this same end-effector-delta $\rightarrow$ velocity-controller interface, differing only in the source of the end-effector pose deltas.

We teleoperate the dual Kinova Gen3 arms using a Meta Quest~3 headset with Quest2ROS~\cite{welle2024quest2ros}, which streams 6-DoF controller poses at approximately 72\,Hz. For each arm, the teleoperation node differentiates consecutive VR poses to obtain end-effector pose deltas, transforms them from the VR frame to the arm base frame, and forwards the resulting twists directly to the velocity controller.

At deployment, the VLA policy outputs sequences of absolute end-effector pose targets. For each arm, we compute the delta to the current pose, convert it into a Cartesian twist via a simple proportional (P) gain, and execute it through the same Kinova velocity controller.

\begin{figure}[t]
  \centering
  \footnotesize
  \setlength{\abovecaptionskip}{0.1cm}
  \includegraphics[width=1\linewidth]{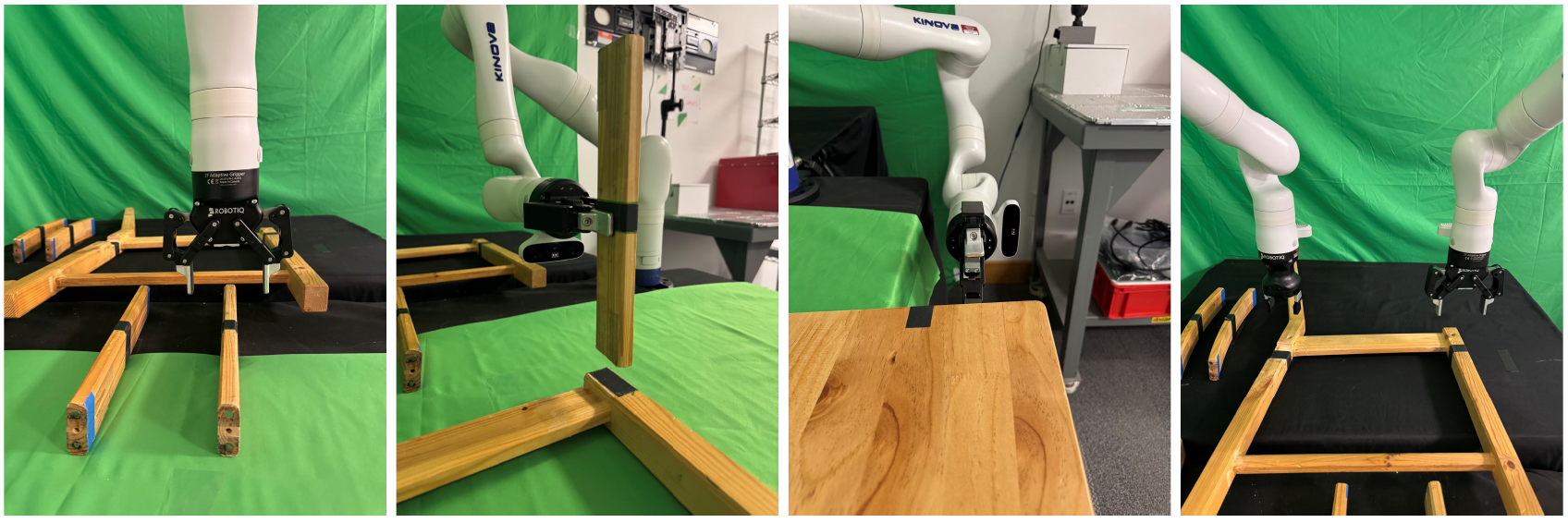}
  \vspace{-3mm} \caption{\textbf{Predefined orientation presets.} Left to right: grasp poses for rail grasping, rail attachment, seat panel grasping, and dual-arm coordinated grasping of the left side frame.}
  \label{fig:grasp}
  \vspace{-1.0em}
\end{figure}

\vspace{0.5em}
\noindent\textbf{Teleoperation Control Mapping.} Since the teleoperator stands opposite the dual-arm system, control is defined in the teleoperator’s frame rather than the robot’s egocentric frame. Motions are mirrored such that directional commands remain intuitive (e.g., moving left/right or forward/backward results in the robot moving in the same perceived direction from the teleoperator’s viewpoint). Controller button mappings are as follows (identical on both hands unless noted):

\vspace{0.5em}
\begin{itemize}
\renewcommand\labelitemi{\scalebox{0.75}{$\bullet$}}
\item \textbf{Index trigger} — engage translation-only control (orientation frozen); releasing freezes the arm (clutching).
\item \textbf{Middle trigger} — engage rotation-only control (position frozen).
\item \textbf{Both triggers held} — full 6-DoF velocity control.
\item \textbf{Lower face button} (A on right, X on left) — toggle the corresponding gripper open/closed.
\item \textbf{B button} (right) — emergency stop on both arms; \textbf{Y button} (left) — trigger a dual-arm coordinated bring-to-pose preset (Fig.~\ref{fig:grasp}).
\item \textbf{Left thumbstick up} — home both arms to the simulator-matched reset pose (interruptible only by B/Y).
\item \textbf{Left/right/down thumbstick} — snap the end-effector to a predefined orientation preset while holding its current position (Fig.~\ref{fig:grasp}).
\item \textbf{Both index triggers held simultaneously} — translation sync mode: the left arm mirrors the right arm's translation.
\item \textbf{Both middle triggers held simultaneously} — rotation sync mode: the left arm mirrors the right arm's rotation (combining both yields fully mirrored 6-DoF motion).
\end{itemize}
\vspace{0.5em}

\vspace{0.5em}
\noindent\textbf{Observations.}
Fig.~\ref{fig:obv_example} shows rear and wrist camera images captured by the system.

\vspace{0.5em}
\noindent\textbf{Furniture Setup and Initialization.}
We use small neodymium (NdFeB) magnets. The magnets are cylindrical, with a diameter of approximately 6\,mm and thickness of 2\,mm, and provide sufficient holding force to stabilize parts upon contact. Magnets are attached inside screwing holes using epoxy glue. Colored markers (e.g., black and blue tape) are placed on parts to assist the teleoperator with visual alignment during demonstration collection. Fig.~\ref{fig:ivar_init} shows example initializations for the LACK side table and IVAR chair assembly tasks.

\vspace{0.5em}
\noindent\textbf{Demonstration Processing.}
Raw demonstrations are recorded at 10\,Hz. We filter no-op actions following DROID~\cite{KhazatskyP0BDKN24}, which reduces the average duration per demonstration from 3\,min to 2.5\,min for the IVAR chair, but do not trim the initial and trailing frames of each non-idle segment, as they contain critical signals indicating subtask completion. Training, inference, and parameters follow simulation (Sec.~\ref{sec:Simulation Implementation Details}).

\vspace{0.5em}
\noindent\textbf{Progress Labels for Real Demonstrations.}
Real demonstrations use the same subtask decomposition and skill primitives as simulation (Table~\ref{tab:skills}). A human annotator marks the primitive boundaries in each recorded demonstration, and progress is then computed from these boundaries with Eq.~\ref{eq:3}, exactly as in simulation.

\begin{table*}[t]
\setlength{\abovecaptionskip}{0.1cm}
\footnotesize
\caption{\textbf{Finetuning hyperparameters of FurnitureVLA.}}
\begin{adjustbox}{width=1.0\linewidth,center}
\centering
\begin{tabular}{@{}p{3.5cm}p{12cm}@{}}
\toprule
\textbf{Hyperparameter} & \textbf{Value} \\
\noalign{\vskip 0.3ex}
\hline
\noalign{\vskip 1.0ex}
pretrained checkpoint & \texttt{pi05\_base} \\
\noalign{\vskip 0.5ex}

\# GPUs & 8 $\times$ NVIDIA L40S (48GB VRAM) \\
\noalign{\vskip 0.5ex}

learning rate (LR) & $2.5\times10^{-5}$ (cosine decay to $2.5\times10^{-6}$ over 30K steps, with 1K-step linear warmup) \\
\noalign{\vskip 0.5ex}

optimizer & AdamW ($\beta_1{=}0.9$, $\beta_2{=}0.95$, $\epsilon{=}10^{-8}$, weight decay $10^{-10}$, grad-norm clip $1.0$) \\
\noalign{\vskip 0.5ex}

effective batch size & 64 (8 per GPU, no gradient accumulation) \\
\noalign{\vskip 0.5ex}

\# train steps & 40K \\
\noalign{\vskip 0.5ex}

\# denoising steps (inference) & 10 (flow-matching) \\
\noalign{\vskip 0.5ex}

input images & 4 RGB views: 1 front, 2 wrist-mounted, 1 rear \\
\noalign{\vskip 0.5ex}

input image size & 448 $\times$ 448 \\
\noalign{\vskip 0.5ex}

use observation history & no (single-step inputs) \\
\noalign{\vskip 0.5ex}

action chunk size ($H$) & 50 steps \\
\noalign{\vskip 0.5ex}

action dimensions & 15 (14 robot + 1 progress) \\
\noalign{\vskip 0.5ex}

use proprio (robot state) & yes (discretized and inlined into the language prompt) \\
\noalign{\vskip 0.5ex}


use LoRA & no (full finetune) \\
\noalign{\vskip 0.5ex}

\# trainable parameters & $\sim$2.6B total (2.3B PaliGemma-2B backbone + 311M Gemma-300M action expert + $<$1M action/state projection heads) \\
\noalign{\vskip 0.5ex}

image augmentations & resize-with-pad \\

\bottomrule
\end{tabular}
\label{tab:hyperparameters_training}
\end{adjustbox}
\vspace{-0.5em}
\end{table*}

\section{Additional Details of FurnitureVLA}\label{sec:FurnitureVLA Details}
\noindent\textbf{VLA Preliminaries.}
A VLA policy $\pi_\theta$ maps an observation $o_t$ (RGB images and proprioception) and a language instruction $g$ to robot actions. We represent actions in a continuous absolute end-effector space. For bimanual control, actions for the left and right arms are defined as $a_t^{L}, a_t^{R} \in \mathbb{R}^7$, where each action is defined as $[x_t, y_t, z_t, u_t, v_t, w_t, \gamma_t]^\top$, representing the target end-effector pose and gripper state. The full action is formed by concatenation, $a_t \in \mathbb{R}^{14}$. The policy uses flow matching~\cite{pi0_5_pi} to decode a chunk of $H$ future actions in a single forward pass: $\pi_\theta(a_{t:t+H-1} \mid o_t, g)$.

\vspace{0.5em}
\noindent\textbf{Training Hyperparameters.} We use $\pi_{0.5}$~\cite{pi0_5_pi} as the VLA backbone, finetuned for 40{,}000 steps on 8 NVIDIA L40S GPUs with global batch size 64. Inference runs on a single L40S GPU with $\tau_p = 0.95$. Table~\ref{tab:hyperparameters_training} lists the hyperparameters used to finetune FurnitureVLA on the $\pi_{0.5}$ backbone. We perform full finetuning.

\vspace{0.5em}
\noindent\textbf{Resolution Upscaling.}
To retain fine-grained visual cues for contact-rich bimanual assembly, we increase the input resolution of the PaliGemma--SigLIP vision backbone from its pretrained $224{\times}224$ to $448{\times}448$. The target resolution aligns with the SigLIP So400m/14 patch size, yielding a $32{\times}32$ token grid ($448/14=32$) without truncation. Input images are resized with aspect-ratio-preserving interpolation and zero-padded to $448{\times}448$. The pretrained positional embeddings ($24{\times}24$) are resized to match the new $32{\times}32$ token grid via bicubic interpolation. All other pretrained parameters are kept unchanged, and the backbone is finetuned end-to-end with the action expert.

\vspace{0.5em}
\noindent\textbf{Temporal Ensembling.}
Following ACT~\cite{aloha23}, we maintain a rolling buffer of $B$ overlapping action chunks and execute a weighted average of all predictions for the current timestep, excluding the progress dimension, where $a_{t}^{[t-i]} \in \mathbb{R}^{14}$ is the prediction for timestep $t$ made $i$ steps ago:
\begin{equation}
\hat{a}_t = \frac{\sum_{i=0}^{B-1} w_i \, a_{t}^{[t-i]}}{\sum_{i=0}^{B-1} w_i},
\quad w_i = e^{\lambda i}.
\label{eq:5}
\end{equation}
More negative $\lambda$ emphasizes recent predictions, while $\lambda{=}0$ gives uniform averaging. The progress signal $\hat{p}_t$ always uses the most recent prediction, independent of the ensembling buffer.

\begin{table}[t]
\setlength{\abovecaptionskip}{0.1cm}
    \footnotesize
    \caption{\textbf{Handover states of the Fabrica-style hybrid.} The planner hands control to FurnitureVLA when its state machine first enters the listed state (i.e., when the part is held above its target).}
    \begin{adjustbox}{width=1.0\linewidth,center}
    \centering
    \begin{tabular}{llc}
    \toprule
    \textbf{Furniture} & \textbf{Part(s)} & \textbf{Handover state} \\ \noalign{\vskip 0.3ex}
    \hline \noalign{\vskip 1.0ex}
    LACK & legs & \texttt{reach\_tabletop\_z} \\
    \noalign{\vskip 0.5ex}
    KALLAX & side panels, center divider & \texttt{reach\_bottom\_z} \\
    \noalign{\vskip 0.5ex}
    KALLAX & top panel & \texttt{insert} \\
    \noalign{\vskip 0.5ex}
    IVAR & rails & \texttt{reach\_frame\_z} \\
    \noalign{\vskip 0.5ex}
    IVAR & left side frame & \texttt{lower\_to\_hover} \\
    \noalign{\vskip 0.5ex}
    IVAR & partially assembled chair frame & \texttt{lower\_after\_flip} \\
    \noalign{\vskip 0.5ex}
    IVAR & seat panel & \texttt{right\_translate\_yz} \\
    \bottomrule
    \end{tabular}
    \label{tab:hybrid_handover}
\end{adjustbox}
\vspace{-1.0em}
\end{table}

\begin{table*}[t]
\setlength{\abovecaptionskip}{0.1cm}
    \footnotesize
    \caption{\textbf{Failure predicates} used for the breakdown in Table~\ref{tab:failures}, evaluated on privileged simulator state from the completion of the previous subtask to the end of the episode. Physical causes are tested in the listed order and one is assigned per failed episode; transition errors are recorded for every subtask switch.}
    \label{tab:failure_predicates}
    \centering
    \begin{adjustbox}{width=\linewidth,center}
    \renewcommand{\arraystretch}{1.15}
    \begin{tabular}{llp{9.2cm}}
    \toprule
    \textbf{Category} & \textbf{Failure Mode} & \textbf{Predicate} \\
    \midrule
    \multirow{4}{*}{\textbf{Physical}}
        & Toppled part           & the part falls by more than 5\,cm before it is ever lifted \\
        & Grasp failure          & the gripper never closes within 30\,cm of the part, or the part is never lifted by more than 3\,cm \\
        & Insertion misalignment & the part is brought within 8\,cm of its target pose but never meets the capture tolerance \\
        & Transport drop         & the part falls by more than 3\,cm below its peak height before reaching the target zone \\
    \midrule
    \multirow{2}{*}{\shortstack[l]{\textbf{Transition}\\\textbf{(co-occurring)}}}
        & Premature transition   & the progress head fires before the current part is captured \\
        & Missed transition      & the part is captured but the progress head never fires within the subtask budget \\
    \bottomrule
    \end{tabular}
    \end{adjustbox}
\end{table*}

\begin{table}[t]
\setlength{\abovecaptionskip}{0.1cm}
    \footnotesize
    \caption{\textbf{Offline progress prediction} on 20 held-out scripted demonstrations per furniture that were never used in training. MAE and p95 are the mean and 95th percentile of the absolute error between the predicted progress and the ground-truth label of Eq.~\ref{eq:3}.}
    \begin{adjustbox}{width=0.7\linewidth,center}
    \centering
    \begin{tabular}{lccc}
    \toprule
    \textbf{Furniture} & \textbf{Frames} & \textbf{MAE} & \textbf{p95} \\ \noalign{\vskip 0.3ex}
    \hline \noalign{\vskip 1.0ex}
    LACK & 2{,}005 & 0.003 & 0.007 \\
    \noalign{\vskip 0.5ex}
    KALLAX & 2{,}854 & 0.003 & 0.008 \\
    \noalign{\vskip 0.5ex}
    IVAR & 5{,}304 & 0.004 & 0.013 \\
    \bottomrule
    \end{tabular}
    \label{tab:progress_offline}
\end{adjustbox}
\end{table}

\section{Implementation of Baselines}\label{sec:Baselines}
\noindent\textbf{Learned Policies without a VLA Backbone.}
We implement Diffusion Policy and ACT on top of our simulation codebase, following the original designs. We adapt them to bimanual actions and our four camera views. Both are trained from scratch on the same demonstrations as FurnitureVLA.

\vspace{0.5em}
\noindent\textbf{Motion Planner with Open-Loop Replay.}
The data-collection motion planner is run with ground-truth part poses and its recorded end-effector actions are replayed open-loop from the same initial layout, using the replay procedure that validates our demonstrations (Sec.~\ref{sec:Simulation Data Generation and Success Criteria}). The replay is fully physical, including the bimanual transports, which are executed kinematically during data generation (Sec.~\ref{sec:Simulation Data Generation and Success Criteria}). Its success rate in Table~\ref{tab:long_horizon} is the fraction of planned episodes that survive this replay; only those form the demonstration set.

\vspace{0.5em}
\noindent\textbf{Fabrica-Style Hybrid.}
Following Fabrica~\cite{tian2025fabrica}, the planner handles sequencing, grasping, and transport with ground-truth part poses, and hands control to FurnitureVLA once the part is held above its target (Table~\ref{tab:hybrid_handover}); the VLA then performs the contact-rich insertion with the subtask instruction, and the planner resumes with its retreat and the next part once the privileged assembly predicate holds.

\vspace{0.5em}
\noindent\textbf{Hi Robot-Style VLM and Oracle Transitions.}
Both coordinators keep FurnitureVLA as the low-level controller and replace its progress-triggered transition with the same switch rule: once the current part is judged attached, the coordinator waits the retreat duration seen in the demonstrations and then switches, so that the next subtask starts from the same post-retreat state as in training. They differ only in who judges attachment. The VLM coordinator queries GPT-6 Astra~\cite{GPT-6-astra_openai} at 1\,Hz with the front and rear images, the subtask list, and the current instruction (prompt in Sec.~\ref{sec:Prompt of the Hi Robot-Style Baseline}); the oracle coordinator uses the simulator's assembly predicate.

\section{Additional Analyses}\label{sec:Analyses Details}
\noindent\textbf{Failure Classification.}
Every failed rollout is classified automatically from privileged simulator state with the predicates of Table~\ref{tab:failure_predicates}, assigned at its first uncompleted subtask.

\vspace{0.5em}
\noindent\textbf{Failure Analysis Human Evaluation.}
To validate the automatic labels, a human annotator inspected 21 failed rollouts covering all furniture and categories from the front and rear videos of the failed subtask. The automatic label agreed with the video in 20 of 21 cases. In the remaining case the video could not tell whether the gripper closed on nothing or never closed at all; both are grasp failures, and we report them as one class, at which level all 21 labels agree.

\begin{table}[t]
\setlength{\abovecaptionskip}{0.1cm}
    \footnotesize
    \caption{\textbf{Progress-triggered retry} (success rates~$\uparrow$, 100 rollouts per furniture on identical layouts, model finetuned with 50\% of the demonstrations). Triggered: episodes in which the stall trigger fired; recovered: triggered episodes that succeed after retrying.}
    \begin{adjustbox}{width=1.0\linewidth,center}
    \centering
    \begin{tabular}{lcccc}
    \toprule
    \textbf{Furniture} & \textbf{No Retry} & \textbf{With Retry} & \textbf{Triggered} & \textbf{Recovered} \\ \noalign{\vskip 0.3ex}
    \hline \noalign{\vskip 1.0ex}
    LACK & 0.82 & 0.87 & 15 & 4 \\
    \noalign{\vskip 0.5ex}
    KALLAX & 0.78 & 0.82 & 18 & 2 \\
    \noalign{\vskip 0.5ex}
    IVAR & 0.44 & 0.46 & 10 & 1 \\
    \noalign{\vskip 0.3ex}
    \midrule \noalign{\vskip 0.2ex}
    \textbf{Average} & \textbf{0.68} & \textbf{0.72} & & \\
    \bottomrule
    \end{tabular}
    \label{tab:retry}
\end{adjustbox}
\end{table}

\vspace{0.5em}
\noindent\textbf{Offline Progress Prediction.}
To measure the accuracy of the progress head independently of the policy's own actions, we collect 20 fresh scripted demonstrations per furniture unseen in training, run the policy on every fifth frame, and compare its predicted progress against the ground-truth label of Eq.~\ref{eq:3} (Table~\ref{tab:progress_offline}). The mean prediction error is well under 1\% of a subtask on all three furniture, which let the $\tau_p{=}0.95$ switch to land reliably inside the demonstrated retreat window.

\vspace{0.5em}
\noindent\textbf{Discrete Progress Supervision.}
As an alternative to the continuous signal, we discretize task-level progress: every observation of subtask $k\in\{1,\ldots,K\}$ is assigned the constant value $(2k-1)/(2K)$, the final frames of a subtask carry the value of the next one so that the head learns to advance, and at inference the active subtask is read from the bin the prediction falls in. This variant fails completely (0.00 on all three furniture): after a subtask completes, the prediction stays in its bin and execution stalls. We attribute this to the visual similarity between nearly assembled and assembled states, which makes the abrupt jump between bins difficult to predict, whereas continuous progress changes smoothly throughout each subtask.

\vspace{0.5em}
\noindent\textbf{Exploratory Progress-Triggered Retry.}
Since the progress estimate stalls when a subtask stops advancing, we test whether it can also drive recovery at inference time, without retraining. To obtain more failures to study, we use the model finetuned with 50\% of the demonstrations (Table~\ref{tab:ablations}). A stall is declared when the predicted progress has not increased by 0.02 for 150 control steps; the policy then opens both grippers, lifts, returns to the end-effector poses recorded at the start of the subtask, and re-issues the same instruction with the progress input and action buffers reset, up to twice per subtask. Table~\ref{tab:retry} shows that the trigger is selective, firing in 43 stalled episodes across the three furniture, but recovery is rare: 7 of the 43 are rescued, and success rates rise by 5, 4, and 2 points. The explanation is in Table~\ref{tab:failures}: the dominant failures leave the scene disturbed, with the part toppled or displaced, and the demonstrations contain no re-grasp from such a state, so a second attempt from the subtask's start pose usually repeats the first. Detecting failure from progress is thus reliable; recovering from it requires recovery behavior in the data, which we leave to future work.

\section{Additional Results}\label{sec:Additional Results}
\noindent\textbf{Simulation Results Breakdown.} We visualize subtask success rates from Table~\ref{tab:design_factors} and Table~\ref{tab:ablations} in Fig.~\ref{fig:appendix_results}.

\begin{figure*}[t]
  \centering
  \footnotesize
  \setlength{\abovecaptionskip}{0.1cm}
  \includegraphics[width=0.70\linewidth]{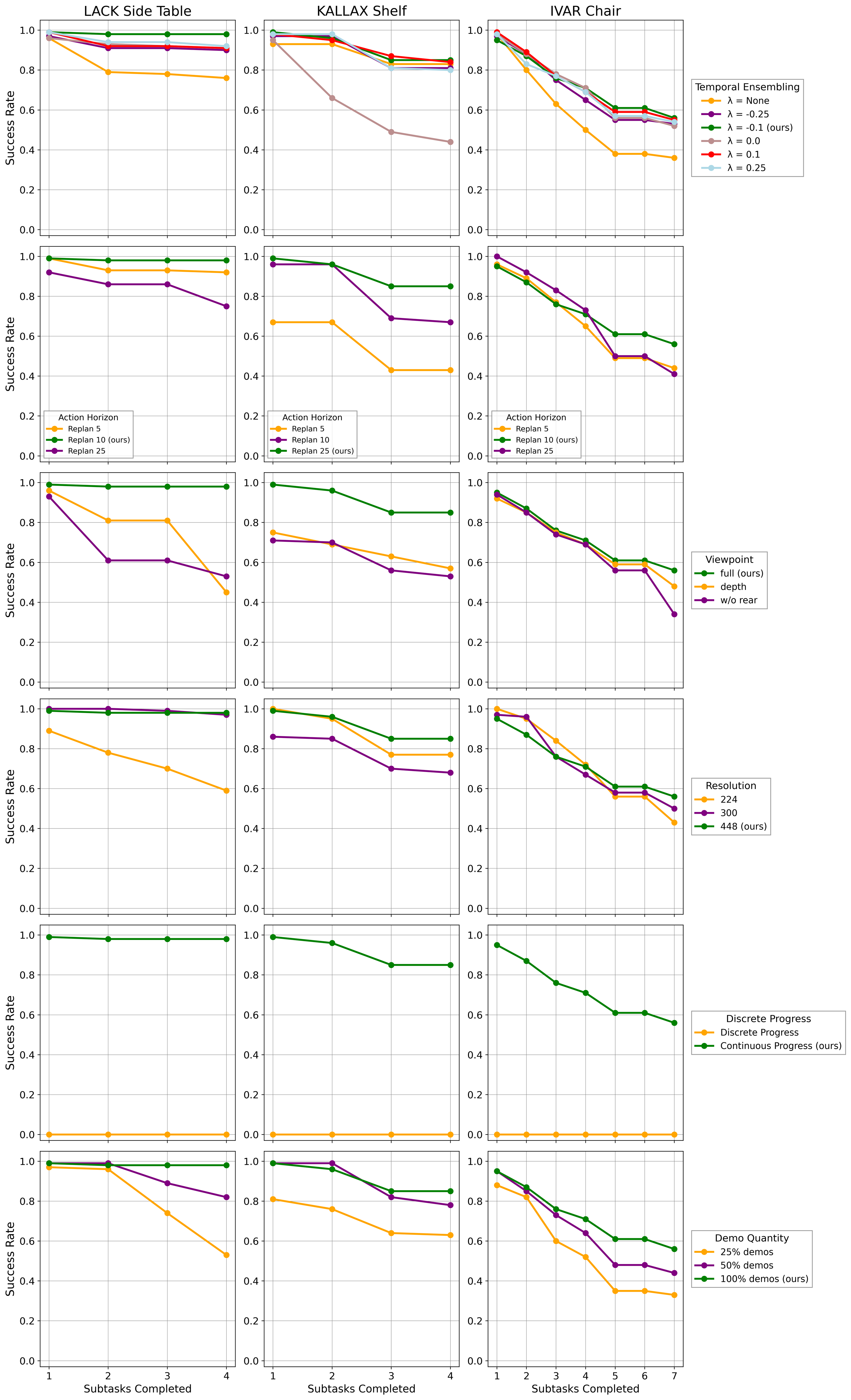}
  \vspace{2mm} \caption{\textbf{Subtask success rates under different design choices.} Results correspond to Table~\ref{tab:design_factors} and Table~\ref{tab:ablations}.}
  \label{fig:appendix_results}
  \vspace{-1.0em}
\end{figure*}

\clearpage
\onecolumn
\section{Prompt of the Hi Robot-Style Baseline}\label{sec:Prompt of the Hi Robot-Style Baseline}
We show the prompt sent to the VLM coordinator at every query, together with the front and rear images. Curly braces mark the fields filled per query; the two furniture-specific descriptions are written once per furniture and are shown here for the LACK side table.

\begin{tcolorbox}[mypromptstyle]
  \prompttitle{Hi Robot-Style Subtask Completion Prompt.}
\begin{flushleft}
\setlength{\parindent}{0pt}
\setlength{\leftskip}{1em}
\setlength{\rightskip}{1em}
You monitor a two-arm robot assembling an \{furniture\_name\} in simulation. \{assembly\_description\} The assembly has \{num\_subtasks\} subtasks, executed in order:\\
\hspace{1em}1. \{subtask\_1\}\\
\hspace{1em}2. \{subtask\_2\}\\
\hspace{1em}\ldots\\
The robot is currently executing subtask \{k\}: "\{current\_subtask\}".\\
Look at the two camera views (front view, rear view). Is the \{current\_part\} now attached in its final place (\{attached\_description\})? It may still be touched by the gripper; judge only the part's placement. Answer strictly as JSON: \{"completed": true or false, "confidence": number between 0 and 1, "reason": short string\}.\\
\vspace{1em}
Furniture-specific descriptions (LACK side table):\\
\hspace{1em}\parbox[t]{\dimexpr\linewidth-3em}{\hangindent=0.6em\hangafter=1 - assembly\_description: The tabletop lies upside down on the work table and the four legs are inserted one by one into its corners, so an attached leg stands vertically on a corner of the tabletop, pointing up.}\\[0.4em]
\hspace{1em}\parbox[t]{\dimexpr\linewidth-3em}{\hangindent=0.6em\hangafter=1 - attached\_description: standing vertically on its corner of the upside-down tabletop}
\end{flushleft}
\end{tcolorbox}

\fi

\end{document}